\documentclass[10pt,journal,compsoc]{IEEEtran}

\ifCLASSOPTIONcompsoc
  \usepackage[nocompress]{cite}
\else
  \usepackage{cite}
\fi

\usepackage{amsmath,amsfonts,bm}

\def\eqref#1{equation~\ref{#1}}

\def\1{\bm{1}}

\def\vn{{\bm{n}}}

\def\vp{{\bm{p}}}
\def\vq{{\bm{q}}}

\DeclareMathAlphabet{\mathsfit}{\encodingdefault}{\sfdefault}{m}{sl}
\SetMathAlphabet{\mathsfit}{bold}{\encodingdefault}{\sfdefault}{bx}{n}

\usepackage{color,url}

\usepackage{multirow}
\usepackage{makecell}
\usepackage{subcaption}
\usepackage{booktabs}
\usepackage{bm}
\usepackage{amsmath,amssymb}
\usepackage{graphicx}
\usepackage{stfloats}

\usepackage[pagebackref=true,breaklinks=true,colorlinks,bookmarks=false]{hyperref}

\usepackage{placeins}

\begin{document}

\title{Geometry-Aware Generation of Adversarial Point Clouds}

\author{
    Yuxin Wen, 
    \and Jiehong Lin, 
    \and Ke Chen, 
    \and C. L. Philip Chen, ~\IEEEmembership{Fellow,~IEEE}, 
    \and and Kui Jia
    \IEEEcompsocitemizethanks{
    \IEEEcompsocthanksitem{
        Y. Wen, J. Lin, K. Chen, and K. Jia are with the School of Electronic and Information Engineering, South China University of Technology, Guangzhou 510641, China, and also with Pazhou Lab, Guangzhou 510335, China, and with Peng Cheng Laboratory, Shenzhen 518005, China. E-mail: \{wen.yuxin, lin.jiehong\}@mail.scut.edu.cn, \{chenk, kuijia\}@scut.edu.cn.}
    \IEEEcompsocthanksitem{
        C. L. Philip Chen is with the School of Computer Science and Engineering, South China University of Technology, Guangzhou 510641, China, and also with Pazhou Lab, Guangzhou 510335, China. E-mail: philip.chen@ieee.org.}
    \IEEEcompsocthanksitem{Corresponding author: Kui Jia.}
    }    
}

\IEEEtitleabstractindextext{

\begin{abstract}
Machine learning models have been shown to be vulnerable to adversarial examples.
While most of the existing methods for adversarial attack and defense work on the 2D image domain, a few recent attempts have been made to extend them to 3D point cloud data.
However, adversarial results obtained by these methods typically contain point outliers, which are both noticeable and easy to defend against using the simple techniques of outlier removal.
Motivated by the different mechanisms by which humans perceive 2D images and 3D shapes, in this paper we propose the new design of \emph{geometry-aware objectives}, whose solutions favor (the discrete versions of) the desired surface properties of smoothness and fairness.
To generate adversarial point clouds, we use a targeted attack misclassification loss that supports continuous pursuit of increasingly malicious signals.
Regularizing the targeted attack loss with our proposed geometry-aware objectives results in our proposed method, \emph{Geometry-Aware Adversarial Attack ($GeoA^3$)}.
The results of $GeoA^3$ tend to be more harmful, arguably harder to defend against, and of the key adversarial characterization of being imperceptible to humans.
While the main focus of this paper is to learn to generate adversarial point clouds, we also present a simple but effective algorithm termed $Geo_{+}A^3$-IterNormPro, with Iterative Normal Projection (IterNorPro) that solves a new objective function $Geo_{+}A^3$, towards surface-level adversarial attacks via generation of adversarial point clouds.
We quantitatively evaluate our methods on both synthetic and physical objects in terms of attack success rate and geometric regularity.
For a qualitative evaluation, we conduct subjective studies by collecting human preferences from Amazon Mechanical Turk.
Comparative results in comprehensive experiments confirm the advantages of our proposed methods.
Our source codes are publicly available at \href{url}{https://github.com/Yuxin-Wen/GeoA3}.
\end{abstract}


}

\maketitle
\IEEEdisplaynontitleabstractindextext
\IEEEpeerreviewmaketitle

\IEEEraisesectionheading{\section{Introduction}\label{sec:introduction}}

\IEEEPARstart{M}{odern} machine learning models, particularly those based on deep networks, have achieved remarkable success on a variety of semantic tasks, with the classification of 2D images \cite{he2015deep,huang2016densely,jia2019orthogonal} and 3D point clouds \cite{PointNet,PointNetPP,DGCNN,tang2020improving} as prominent examples.
In spite of these successes, these models are vulnerable to \textit{adversarial examples} \cite{szegedy2013intriguing} --- specially crafted perturbations of input data that are imperceptible to humans but that would cause failure of these classification models.
This phenomenon prevails across models and data types \cite{su2018robustness,fawzi2018adversarial,CW3DAdv}.
The existence of adversarial examples has triggered a large amount of research focusing either on attack/defense studies for safety-critical applications \cite{sharif2016accessorize,cao2019adversarial,wiyatno2019physical}, or on the robustness analysis of machine learning models \cite{ilyas2019adversarial,zhang2019theoretically,schmidt2018adversarially}.

The seminal work of Szegedy \emph{et al.} \cite{szegedy2013intriguing} discovered adversarial 2D images by analyzing the classification robustness of deep networks.
Inspired by \cite{szegedy2013intriguing}, many works have been proposed for adversarial 2D images, where most of them played attack-and-defense games that a successful defense against the previous attack would always be breached by the subsequent attack.
In various algorithms \cite{goodfellow2014explaining,CW,PGD}, the most important characterization of the produced adversarial images was always \emph{imperceptibility}.
To achieve imperceptibility, these methods typically constrain the $l_p$-norms of adversarial noise to be small on the 2D image domain.
The effectiveness of this scheme is grounded on how humans perceive 2D images.
Indeed, when adding noise of small magnitude to a benign image, human perception is overwhelmed by the apparent patterns contained in the original image, rather than by the added high-frequency but low-magnitude noise that has no semantic patterns; adversarial imperceptibility is thus achieved.

\begin{figure*}[htbp]
	\vskip 0.15in
	\begin{center}
		\includegraphics[width=0.32\textwidth]{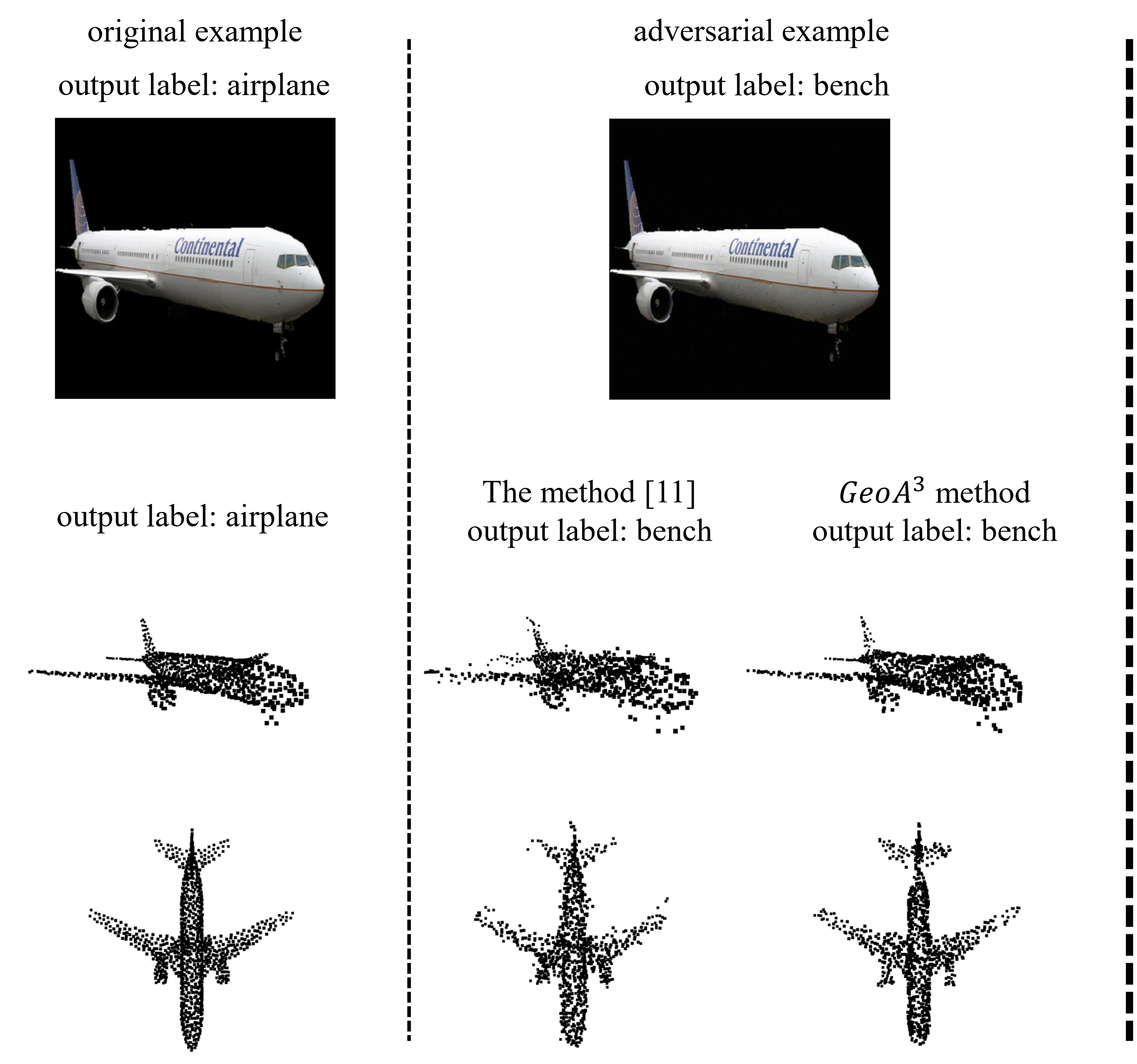}
		\includegraphics[width=0.32\textwidth]{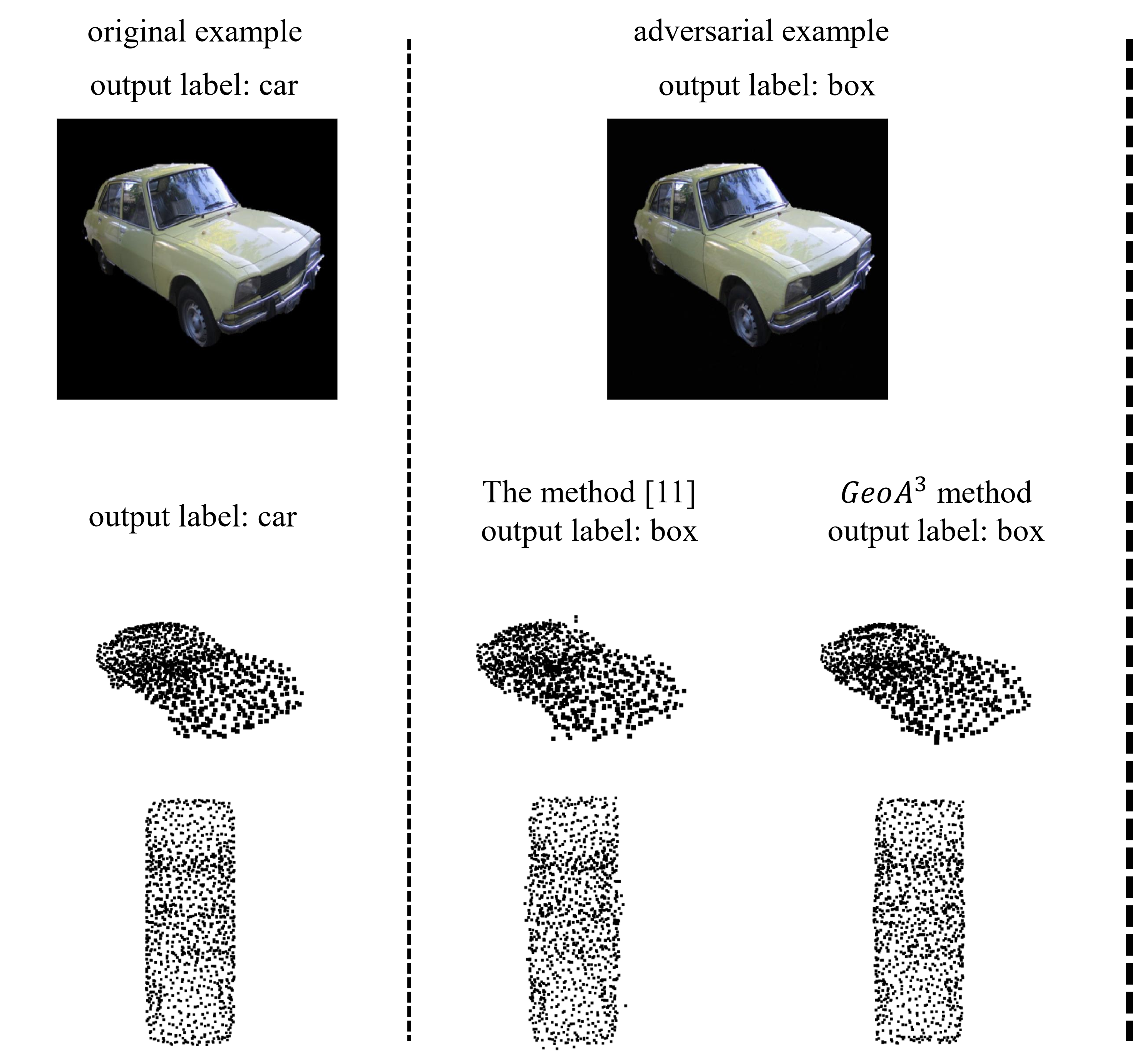}
		\includegraphics[width=0.32\textwidth]{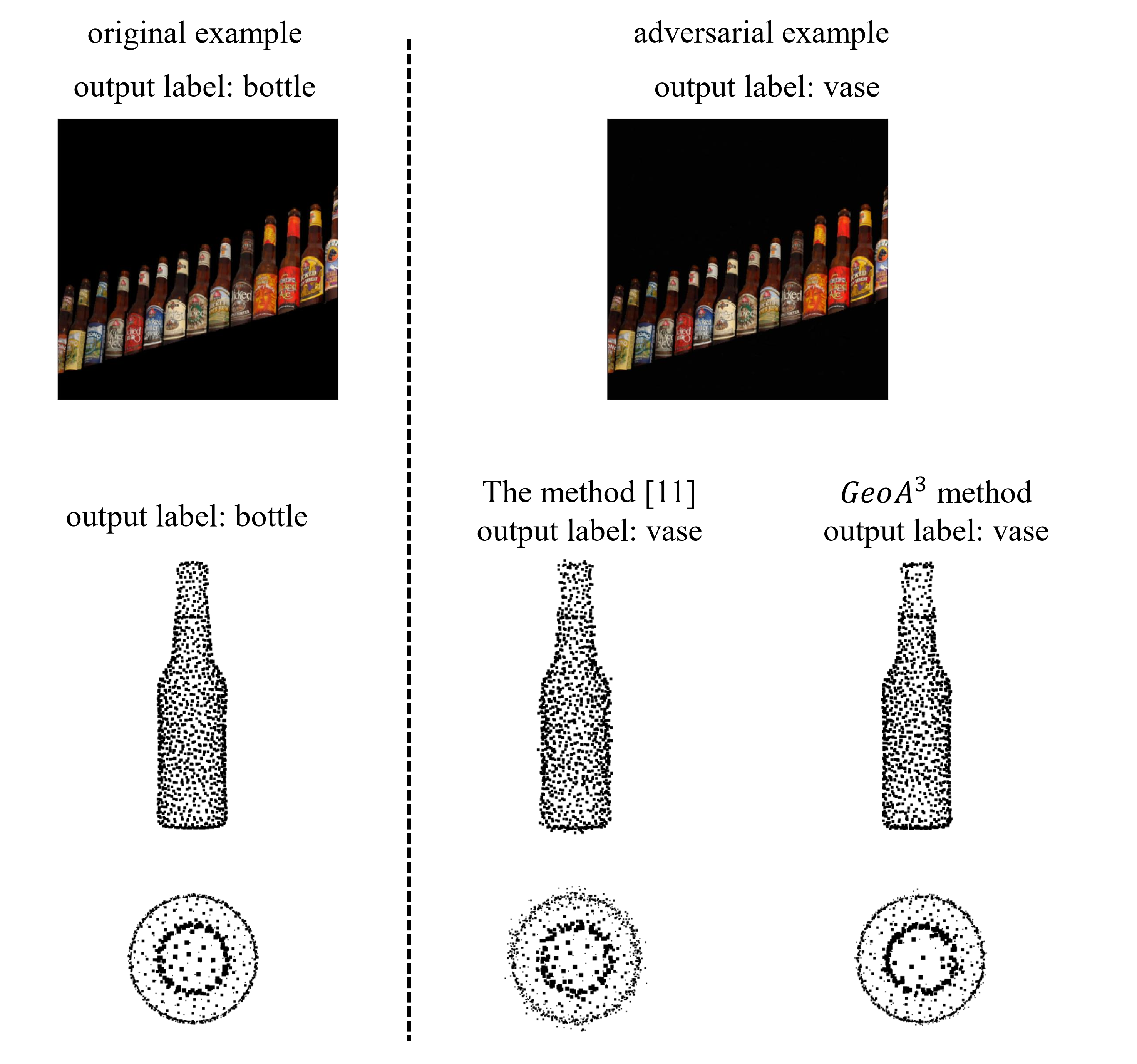}
	\end{center}
	\caption{Adversarial examples of 2D images and 3D point clouds corresponding to the same object categories from the PASCAL3D+ dataset \cite{xiang2014beyond}. Adversarial images are obtained by using C\&W attack \cite{CW} against the Inception-V3 \cite{szegedy2016rethinking} model of image classification. Adversarial point clouds are obtained respectively by using the method \cite{CW3DAdv} and our proposed $GeoA^3$ against PointNet model \cite{PointNet}.}\label{fig:2D3DAdvPerceptionDiff}
	\vskip -0.1in
\end{figure*}

More recently, this general idea of adversarial attack has been employed in 3D point cloud data \cite{CW3DAdv,Extending,yang2019adversarial,liu2019adversarial}.
These studies learned to either perturb individual points contained in a point cloud or to attach additional points to it, causing the resulting point cloud to be misclassified by a point set classifier of interest \cite{PointNet,PointNetPP,DGCNN}.
To achieve imperceptibility, these methods followed other pieces of research on adversarial 2D images and constrained their adversarial point clouds to be close to benign ones under certain distance metrics of point set.
This seems like a straightforward technical extension at first glance.
Unfortunately, the adversarial point clouds obtained by these methods typically contain point outliers, and these point outliers are particularly noticeable when humans perceive the underlying surface represented by the point cloud, as shown in Figure \ref{fig:2D3DAdvPerceptionDiff}.
In this work, we analyze in Section \ref{SecAdvPCNature} that there exists a sharp difference between the human perception of 2D images and that of 3D shapes, especially approximate surface representations of 3D shapes such as point cloud data that contain no texture and shading information.
Existing methods \cite{CW3DAdv,yang2019adversarial} did not take this into consideration and thus caused a ``spiky'' perception of the surfaces boundaries and thus draw humans' attention.
In addition, these methods produce adversarial point clouds that tend to be defended easily through the removal of the generated point outliers via simple statistical techniques \cite{Deflecting}.

To address these issues, we are motivated by the fact that an adversarial point cloud, as a discrete approximation of an object's surface, should satisfy (the discrete versions of) the general surface properties of \emph{smoothness} and \emph{fairness} \cite{botsch2010polygon}.
These properties are related to the continuity and variation of the (partial) derivatives of a parametric surface function, and in particular, to the curvatures of local surface patches.
To achieve imperceptibility, we propose the new solution of \emph{geometry-aware objectives}, whose design causes the adversarial point cloud to bear the aforementioned general surface properties while also being close to the benign point cloud under distance metrics of point set as a whole.
Technically, our proposed geometry-aware objectives enhance the classical distance terms of point set with a term that promotes the consistency of local curvatures between adversarial and benign point clouds.
In this work, we consider the setting of targeted attack \cite{CW,CW3DAdv} and use a misclassification loss that is able to achieve higher maliciousness when compared with traditional margin-based approaches \cite{CW,CW3DAdv}.
Regularizing the more aggressive misclassification loss with our proposed geometry-aware objectives results in our proposed method, \emph{Geometry-Aware Adversarial Attack ($GeoA^3$)}, which is expected to produce more harmful point clouds that are arguably harder to defend against, without introducing noticeable modifications.

The main focus of this paper is to learn how to generate adversarial point clouds.
This would cause safety-critical issues practically by guiding, for example, a LiDAR spoofer \cite{LiDARSpoofer} to fool a 3D sensor.
Nonetheless, it is more desirable to make the underlying surfaces represented by point clouds adversarial; otherwise, a benign surface re-sampling could possibly defend against such an adversarial attack.
As an attempt to generate adversarial surfaces via the generation of adversarial point clouds, in this work we present a new objective function $Geo_{+}A^3$ and a simple but effective solving algorithm termed \emph{Iterative Normal Projection (IterNormPro)}.
When optimizing the objective of $Geo_{+}A^3$, IterNormPro introduces point-wise projection of 3D coordinates onto the normal direction.
By accumulating iteratively, IterNorPro pursues adversarial update optimization directions that account more for surface deformation.

We quantitatively evaluate adversarial point clouds generated by different methods in terms of attack success rate, under the state-of-the-art defense \cite{Deflecting} and geometric regularity.
For evaluation, we also propose a new measure that is based on the same desired low curvature surface property.
To be more specific, the measurement (cf. equation (\ref{Eq:GeoRegularityMeasure})) is an approximate surrogate of the globally maximum curvature defined on the discrete point cloud.
Such a measurement is relevant to the measure of perceptual aesthetics \cite{MiuraAesthetics,LigangAesthetics}, since both of them are motivated from the same desired surface property of fairness.
For a qualitative evaluation, we conduct subjective studies by collecting human preferences from Amazon Mechanical Turk.
An evaluation of surface-level adversarial effects is conducted by first reconstructing object meshes from the obtained adversarial point clouds and then evaluating the remaining adversarial effects via a re-sampling of the point clouds from the reconstructed mesh surfaces.
A physical attack is similarly conducted by 3D printing the mesh reconstructions and then re-scanning the printed objects for the point clouds to be evaluated.
We present comprehensive experiments that show the advantages of our proposed methods over existing ones.

\subsection{Related Works}\label{SecRelatedWorks}

The existing studies that are closely related to the present study can be organized into the following three lines of research.

\vspace{0.1cm}\noindent\textbf{Adversarial Attack of 2D Images --}
The existence of adversarial examples for deep image classification networks was first suggested in \cite{szegedy2013intriguing}.
Subsequent research proposed various methods to generate adversarial examples that are less likely to be perceived by humans, but are still capable of attacking classification models.
Representative methods include Fast Gradient Sign Method (FGSM) \cite{goodfellow2014explaining}, C\&W attack \cite{CW} and Projected Gradient Descent (PGD) \cite{PGD}.
These methods typically search for low-magnitude pixel-level noise of on the domain of benign images by optimizing misclassification loss functions.
After which, by superimposing the obtained noise onto the respective benign ones, adversarial results are produced.
More specifically, Goodfellow \emph{et al.} \cite{goodfellow2014explaining} found that by moving a benign image towards the decision boundary of a classification model via a single gradient update step, an adversarial image could be generated.
The C\&W attack proposed in \cite{CW} includes both the misclassification loss and a loss constraining the magnitudes of adversarial noise into the optimization objective.
Empirical results show that adversarial images generated by the C\&W attack are both aggressive and difficult to be perceived.
PGD \cite{PGD} is a multi-step optimization method that can be regarded as an improved version of FGSM.
These methods adopt $l_p$-norms to constrain the magnitudes of adversarial noise, which has been shown to effectively satisfy the adversarial criterion of imperceptibility.
Apart from $l_p$-norms, other measures such as the Perceptual Adversarial Similarity Score (PASS) \cite{rozsa2016adversarial} have been proposed to account more for the mechanisms by which humans perceive 2D images, e.g. by measuring regional illumination, contrast, and structural similarity between adversarial images and benign ones.
Unfortunately, due to the difference in the human perception between 2D images and 3D shapes (cf. the discussion in Section \ref{SecAdvPCNature}), both $l_p$-norms and PASS are less relevant for the generation of adversarial point clouds that are expected to be imperceptible to humans.

\vspace{0.1cm}\noindent\textbf{Adversarial Generation of 3D Point Clouds --}
Following these studies on adversarial 2D images, there has been a recent surge of interest in studying the attack of point set classification models \cite{PointNet,PointNetPP,DGCNN} with adversarial point clouds \cite{Extending,CW3DAdv,yang2019adversarial,SalienceMap3D,wicker2019robustness,liu2019adversarial}.
Given a benign point cloud, these methods generated adversarial ones either by perturbing individual points \cite{CW3DAdv,Extending,yang2019adversarial,liu2019adversarial}, by removing some of the points \cite{yang2019adversarial,SalienceMap3D,wicker2019robustness}, or by attaching additional points to the given point cloud \cite{CW3DAdv,yang2019adversarial}.
Intuitively, the strengths of these adversarial attacks depend on the numbers of points that are free to be adjusted via optimization during the generation process.
Consequently, point perturbation is arguably the most effective approach, whereas point detachment is the least effective one --- indeed, point detachment could be more of a way to study the point-wise saliency for classification \cite{SalienceMap3D}.
Technically, existing methods for generating adversarial point clouds follow their 2D image counterparts; namely, they optimize point-wise coordinate offsets with respect to misclassification loss functions defined on point set classification models, where the offset searched is constrained either by $l_p$-norms or by classical point set distance functions.
For example, Xiang \emph{et al.} \cite{CW3DAdv} considered both the settings of point perturbation and point attachment; for the former setting, they adopted the framework of C\&W attack \cite{CW} and used $l_2$-norm to constrain the point-wise perturbations; for the second setting, they used either Chamfer distance or Hausdorff distance to constrain the generation of additional points.
Liu \emph{et al.} \cite{Extending} followed FGSM \cite{goodfellow2014explaining} by using the $l_2$-norm or its variants as constraints.
Despite the success of these methods in their attacks, however, they tend to generate adversarial point clouds that contain clearly visible outliers.
The generation of point outliers both violates the adversarial criterion of imperceptibility and makes defense against them easier (e.g. via the simple techniques of outlier removal \cite{Deflecting}).
It is worth noting that a simple attempt to achieve imperceptibility is made in \cite{Extending} by iteratively projecting the perturbed points back onto the mesh from which the benign point cloud is sampled; unfortunately, attack success rates fall upon adopting this approach.
A recent study by Tsai \emph{et al.} \cite{AAAICloseRelatedWork} was motivated to address the issue of point outliers.
They proposed putting a perturbation-constraining regularization into the C\&W framework, combining a global Chamfer distance and a local term that encourages the compactness of local neighborhoods in the obtained adversarial point cloud (which is achieved by constraining the average distances among points in each neighborhood).
In addition, they proposed a scheme to guide the gradient update directions of adversarial offsets towards those of mesh-level deformations.
However, the adversarial results generated by \cite{AAAICloseRelatedWork} appear to have rugged surfaces, which are easily identified by humans.
By contrast, the geometry-aware objectives proposed in the present work are intuitive and neat, and they are well grounded on the surface properties of smoothness and fairness.
Comparative experiments show that our results are advantageous over those of \cite{AAAICloseRelatedWork} in terms of both geometric regularity and attack success rate.

\vspace{0.1cm}\noindent\textbf{Deep Learning Point Cloud/Surface Reconstruction --}
Our proposed geometry-aware generation of adversarial point clouds is also related to recent methods that learn deep networks to generate 3D surface shapes in the forms of either point clouds \cite{fan2017point,wang} or meshes \cite{Groueix_2018,wang2018pixel2mesh,TangCVPR2019}. Among these methods, Fan \emph{et al.} \cite{fan2017point} made the first attempt to reconstruct a point cloud object surface from only a single image. AtlasNet \cite{Groueix_2018} and subsequent improvements \cite{wang2018pixel2mesh,TangCVPR2019} extended the learning of object surface meshes generation by learning to deform the vertices of input, initial meshes. While some learning terms defined in \cite{wang2018pixel2mesh,TangCVPR2019} are inspired by smooth surface properties as well, the same source of inspiration would suggest different formulations and learning objectives useful in the contexts of surface reconstruction or the generation of adversarial point clouds, respectively. More discussions on these technical differences are presented in Section \ref{SecGeoAwareObj}.

\subsection{Contributions}

Our technical contributions are summarized as follows.

\begin{itemize}
\item Motivated by the difference human perception of 3D shapes from that of 2D images, we propose a new method, termed \emph{Geometry-Aware Adversarial Attack ($GeoA^3$)}, for the generation of adversarial point clouds whose differences from benign ones are expected to be imperceptible to humans.
$GeoA^3$ is built on the new design of \emph{geometry-aware objectives}, which includes the terms of Chamfer distance, Hausdorff distance, and consistency of local curvatures (cf. Section \ref{SecGeoAwareObj}).
Such a design enables the resulting adversarial point clouds to bear (the discrete versions of) the general surface properties of smoothness and fairness.
\item The proposed geometry-aware objectives afford more aggressive attack learning, for which we propose a new targeted attack misclassification loss that the supports continuous pursuit of higher-level malicious signals.
Regularizing the proposed targeted attack loss with geometry-aware objectives is expected to produce more harmful point clouds that are arguably harder to defend against, without introducing noticeable modifications.
\item We also present a simple but effective solving algorithm, termed $Geo_{+}A^3$-IterNormPro, to have the desirable surface-level adversarial effects on the generated adversarial point clouds.
$Geo_{+}A^3$-IterNormPro iteratively accumulates optimization directions of adversarial updates that account more for surface deformation.
The expectation is that certain adversarial effects remain after re-sampling point clouds from surface meshes reconstructed from adversarial point clouds.
\item We quantitatively and qualitatively evaluate adversarial point clouds and their remaining adversarial effects on reconstructed mesh models and 3D-printed physical models.
In addition to attack success rate, we propose a new quantitative measure of geometric regularity that is based on the same desired low curvature surface property.
A qualitative evaluation is also conducted by collecting human preferences from Amazon Mechanical Turk. Comparative results in comprehensive experiments confirm the advantages of our proposed methods over existing ones.
\end{itemize}

Data, source codes, and pre-trained models are made public at \href{url}{https://github.com/Yuxin-Wen/GeoA3}.

\section{Adversarial Point Clouds of Object Surface Shapes}\label{SecGeoAwareAdv}

Our problem setting assumes the availability of a collection of point clouds in the input space $\mathcal{X}$. Any $\mathcal{P} \in \mathcal{X}$ is an approximate shape representation of its underlying surface $\mathcal{S}$ of a certain object category, whose label is $y \in \mathcal{Y}$. Each $\mathcal{P}$ contains an orderless set of $n$ points $\{ \vp_i \}_{i=1}^n$, where any $\vp = [x, y, z]^{\top} \in \mathbb{R}^3$ denotes the coordinates in the Euclidean space.
In this work, we focus on machine learning models of 3D point set classification \cite{PointNet,PointNetPP,DGCNN}, which learn a classifier $f: \mathcal{X} \rightarrow \mathcal{Y}$ and expect $f(\mathcal{P}) = y$ for any input $\mathcal{P}$ of the true label $y$.
Given a learned $f$, we aim to obtain from $\mathcal{P}$ an \emph{adversarial} point cloud $\mathcal{P}'$ by perturbing its individual points $\{ \vp_i \}_{i=1}^n$; the obtained $\mathcal{P}'$ would be misclassified by $f$.
This can be technically achieved by solving
\begin{equation}\label{eq:OverallObjective}
\min_{\mathcal{P}'} C_{Mis}(\mathcal{P}') + \beta \cdot C_{Reg}(\mathcal{P}', \mathcal{P}) ,
\end{equation}
or equivalently
\begin{equation}\label{eq:OverallObjectiveAlt}
\min_{\mathcal{P}'} C_{Mis}(\mathcal{P}') \ \ \mathrm{s.t.} \ \ D(\mathcal{P}', \mathcal{P}) \leq \epsilon ,
\end{equation}
where $C_{Mis}(\mathcal{P}')$ is a loss term defined on $f$ to promote the misclassification of $\mathcal{P}'$, $C_{Reg}(\mathcal{P}', \mathcal{P})$ is a regularizer that penalizes the large difference between $\mathcal{P}$ and the resulting $\mathcal{P}'$, with $\beta$ as the penalty parameter, $D(\mathcal{P}', \mathcal{P})$ is a distance metric, and $\epsilon$ is a small constant.
Similar to adversarial examples of 2D images \cite{szegedy2013intriguing}, the adversarial objective (\ref{eq:OverallObjective}) or (\ref{eq:OverallObjectiveAlt}) suggests that $\mathcal{P}'$ should be close to $\mathcal{P}$ under certain distance metrics, such that the change from $\mathcal{P}$ to $\mathcal{P}'$ is \emph{imperceptible} to humans. For example, the seminal work of Xiang \emph{et al.} \cite{CW3DAdv} implements the constraint of (\ref{eq:OverallObjectiveAlt}) simply as $ \|\bm{p}_i' - \bm{p}_i\|_2 \leq \epsilon \ \forall \ i \in \{1, \dots, n\}$; in other words, it uses $l_2$-norm that constrains the perturbations of individual points in $\mathcal{P}$ to be small \footnote{We note that there exist two strategies in Xiang's work \cite{CW3DAdv} for generation of adversarial point clouds, namely, \emph{point perturbation} and \emph{(additional) point generation}.
The first one does so by learning to perturb individual points contained in an input, benign point cloud, where $l_2$-norm is used to constrain the perturbations of individual points.
The second one learns to generate additional points, and then attaches them to the benign point cloud to form the adversarial result, where either Chamfer distance or Hausdorff distance (cf. Section \ref{SecGeoAwareObj} for their definitions) are used to constrain the generation of additional points.}.

\subsection{The (Im)perceptibility of Adversarial Point Clouds}\label{SecAdvPCNature}

Studies on adversarial examples stem from robust analysis of deep networks, particularly for convolutional networks that are trained to classify 2D images.
It is discovered in \cite{szegedy2013intriguing} that for images correctly classified by a network, superimposing certain noise of small magnitude to them can fool the same network; these images are adversarial since magnitudes of the noise are so small such that they are less likely to be perceived by humans.
Most of subsequent research \cite{goodfellow2014explaining,kurakin2016adversarial,CW,PGD} formalizes this problem by proposing various algorithms to search \emph{pixel-wise, independent} noise under the constraints of small $l_p$-norms on the image domain.
With the constraints, appearance patterns of the resulting images are still dominated by the original ones.

While our intended study on adversarial point clouds generally follows the same methodology, there exists a sharp difference between our human perception of 2D images and that of 3D shapes; psychophysical evidence shows that humans perceive object surface shapes from combined variables of motion, texture, shading and boundary contours, among others \cite{FechnerInfoShapePercep}.
Indeed, for adversarial images, the high-frequency information provided by the added pixel-wise, independent noise is overwhelmed by the appearance patterns contained in the original, clean images; consequently, the noise addition is imperceptible to humans.
As approximate surface representations, point cloud data contain no texture and shading, and our shape perception of point clouds is mostly from their point-wise depths and boundary contours.
In fact, studies have shown that contours are the most important characteristic for visually identifying surface shapes, and adding additional surface information of shading, texture, and motion gives only small improvements in shape judgment \cite{Contour4Shape1,Contour4Shape2}.
Consequently, point perturbations would possibly produce point outliers, which caused a ``spiky'' perception of the surfaces boundaries and thus draw humans' attention.
Our sensitivity to spiky surface is also supported from study on the aesthetics of object shapes; in \cite{AestheticsSurvey}, experiments find that when visiting an art gallery, visitors prefer shapes with gentle curves over those with sharp points.
To have a more intuitive understanding of the difference between human perception of 2D images and 3D shapes, we show in Figure \ref{fig:2D3DAdvPerceptionDiff} adversarial examples of 2D images and 3D point clouds corresponding to the same object categories from the PASCAL3D+ dataset \cite{xiang2014beyond}. We can hardly tell from Figure \ref{fig:2D3DAdvPerceptionDiff} the difference between adversarial images and benign ones, but rather easily perceive the manipulation of adversarial point clouds obtained by a representative existing method \cite{CW3DAdv}.
In contrast, our proposed $GeoA^3$ largely improves the imperceptibility, while still being able to attack classification successfully --- in fact, we will show in Section \ref{SecExps} that our method achieves even higher attack success rates than \cite{CW3DAdv}.
We note that the above surface perception issue is largely overlooked in existing works of adversarial point clouds \cite{CW3DAdv,yang2019adversarial}.
In addition to this issue, outlier points to an object surface are also easier to be removed, causing the generated adversaries to be easily defended against \cite{Deflecting}.

As mentioned above, we argue that if adversarial point cloud satisfies the surface properties as that of the general surface, \textit{e.g.}, smoothness and fairness, it's expected to be imperceptible to humans.
According to this, an adversarial point cloud $\mathcal{P}'$ should be close to $\mathcal{P}$ when measured under certain distance metrics of point set; otherwise, humans would notice either the global, possibly topological changes of part configuration of the object surface, or those of local surface details.
Our analysis leads to technical solutions of \emph{geometry-aware objectives} to generate adversarial point clouds, as presented shortly.

\subsection{The Proposed Geometry-Aware Objectives}\label{SecGeoAwareObj}

Analysis in Section \ref{SecAdvPCNature} inspires us to modify a point cloud $\mathcal{P}$ to have $\mathcal{P}'$ in a geometry-aware manner. This can be technically achieved by objectives that constrain the magnitudes of modification under distance metrics of point set as a whole, while taking into account the local surface smoothness of the resulting $\mathcal{P}'$, in particular prevention of generating point outliers.
In this section, we present the proposed geometry-aware objectives.
With geometry-aware objectives and an additional misclassification loss, we can generate adversarial point cloud, which is to be presented in Section \ref{SecAdvPCGen}.

\vspace{0.1cm}\noindent\textbf{Chamfer Distance --}
Given two point sets $\mathcal{P}$ and $\mathcal{P}'$ respectively of $n$ and $n'$ points, the Chamfer distance computes
\begin{equation}\label{eq:CD}
	\begin{split}
		C_{\mathtt{Chamfer}}(\mathcal{P}', \mathcal{P}) = \frac{1}{n'} \sum_{\vp'\in \mathcal{P}'}&\min_{\vp \in \mathcal{P}} \|\vp'-\vp\|_2^2
		\\&+ \frac{1}{n} \sum_{\vp \in \mathcal{P}}\min_{\vp' \in \mathcal{P}'}\|\vp-\vp'\|_2^2 ,
	\end{split}
\end{equation}
which is symmetric with respect to $\mathcal{P}$ and $\mathcal{P}'$.
Although the Chamfer distance (\ref{eq:CD}) is not a strict distance metric, since the triangle inequality does not hold, it is popularly used in the recent literature of learning based 3D shape generation \cite{yang2018foldingnet,groueix2018atlasnet,fan2017point}.
It measures the distance between the two point sets by averaging over the individual deviations of any $\vp \in \mathcal{P}$ from $\mathcal{P}'$ and those of any $\vp' \in \mathcal{P}'$ from $\mathcal{P}$.
We note that Chamfer distance is less effective in prevention of outlier points when generating $\mathcal{P}'$ from $\mathcal{P}$, since a small portion of outliers in $\mathcal{P}'$ increases the distance (\ref{eq:CD}) negligibly.
This shortcoming of Chamfer distance motivates us to additionally use the following Hausdorff distance.

\vspace{0.1cm}\noindent\textbf{Hausdorff Distance --}
For the point sets $\mathcal{P}$ and $\mathcal{P}'$, we consider in this work a non-symmetric Hausdorff distance that concerns with the resulting $\mathcal{P}'$ only, which computes
\begin{equation}\label{eq:HD}
	C_{\mathtt{Hausdorff}}(\mathcal{P}', \mathcal{P}) = \max_{\vp'\in \mathcal{P}'}\min_{\vp \in \mathcal{P}}\|\vp'-\vp\|_2^2 .
\end{equation}
As (\ref{eq:HD}) indicates, the Hausdorff distance finds the largest one among the smallest distances of individual $\vp' \in \mathcal{P}'$ from $\mathcal{P}$.
It is thus sensitive in case that outliers are generated in $\mathcal{P}'$.

\vspace{0.1cm}
Computation of distances (\ref{eq:CD}) and (\ref{eq:HD}) involves individual points contained in $\mathcal{P}$ and $\mathcal{P}'$, but the local surface geometries are not the focus of them; consequently, it is possible that less smooth surface change, including generation of spiky points, would visibly appear in the resulting $\mathcal{P}'$, even though $\mathcal{P}'$ could be close to $\mathcal{P}$ when measured by (\ref{eq:CD}) and/or (\ref{eq:HD}), causing failure to achieve the imperceptible objective of adversarial modification. We introduce the following objective to reduce the less smooth surface modification.

\vspace{0.1cm}\noindent\textbf{Consistency of Local Curvatures --}
Our way to achieve imperceptible modification of adversarial point cloud can be generally described as ensuring the \emph{local consistency of curvatures} between the surface of $\mathcal{P}$ and that of $\mathcal{P}'$, where local consistency means that for a spatially closest pair of surface points respectively from $\mathcal{P}$ and $\mathcal{P}'$, their magnitudes of curvature are similar. Since computations in this work are conducted on the discrete point clouds, we rely on the following discrete notions of point-wise curvature.

For any point $\vp' \in \mathcal{P}'$, we find its closest point $\vp \in \mathcal{P}$ by $\vp = \arg\min_{\vp \in \mathcal{P}} \| \vp' - \vp \|_2 $.
There exist local point neighborhoods $\mathcal{N}'_{\vp'} \subset \mathcal{P}'$ and $\mathcal{N}_{\vp} \subset \mathcal{P}$ respectively associated with $\vp'$ and $\vp$, which are obtained in this work by searching $k$ nearest neighbors, suggesting $ | \mathcal{N}'_{\vp'} | = | \mathcal{N}_{\vp} | = k$.
To capture the local geometry of $\mathcal{N}_{\vp}$, we rely on the following discrete notion
\begin{equation}\label{eq:disCurv}
	\kappa(\vp; \mathcal{P}) = \frac{1}{k} \sum_{\vq \in \mathcal{N}_{\vp} } \left| \left< (\vq - \vp)/\|\vq - \vp\|_2, \vn_{\vp} \right> \right| ,
\end{equation}
where $\left<\cdot, \cdot\right>$ denotes inner product, and $\bm{n}_{\vp}$ denotes the unit normal vector of the surface at $\vp$. We follow \cite{hoppe1992surface} and compute $\vn_{\vp}$ from $\mathcal{N}_{\vp}$ as follows: we first generate a $3\times3$ positive semidefinite covariance matrix
\begin{equation}\label{eq:localcovmat}
\bm{C} = \sum_{\vq\in \mathcal{N}_{\vp}} (\vq-\vp)\otimes(\vq-\vp) ,
\end{equation}
where $\otimes$ denotes the outer product operation; we then apply eigen-decomposition to $\bm{C}$, obtaining $\vn_{\vp}$ as the eigenvector corresponding to the smallest eigenvalue of $\bm{C}$; the first two eigenvectors define the surface tangent plane $\mathcal{T}(\mathcal{N}_{\vp})$ at $\vp$.

The term (\ref{eq:disCurv}) intuitively measures the averaged angles between the normal vector and the vector defined by pointing $\vp$ towards each $\vq$ of its neighboring points.
Indeed, since the normal vector $\vn_{\vp}$ is orthogonal to the tangent plane $\mathcal{T}(\mathcal{N}_{\vp})$ of the surface at $\vp$, each inner product in (\ref{eq:disCurv}) characterizes how the normals vary directionally in the local neighborhood $\mathcal{N}_{\vp}$, thus approximately measuring the local, directional curvature, and an average of $| \mathcal{N}_{\vp} |$ inner products in (\ref{eq:disCurv}) approximately measures the local, mean curvature.
We compute $\kappa'(\vp'; \mathcal{P}')$ in the same way as (\ref{eq:disCurv}), with a subtle difference that instead of computing $\vn'_{\vp'}$ from $\mathcal{N}'_{\vp'}$, we directly use $\vn_{\vp}$, \textit{i.e.,} the unit normal vector of the point in $\mathcal{P}$ that is closest to $\vp'$, as a surrogate of $\vn'_{\vp'}$, since normal vectors of $\mathcal{P}$ can be pre-computed and efficiently retrieved during the modification process.

Given $\kappa'(\vp'; \mathcal{P}', \mathcal{P})$ and $\kappa(\vp; \mathcal{P})$, we use the following objective to encourage the consistency of local geometries between any $\vp' \in \mathcal{P}'$ and its closest point $\vp \in \mathcal{P}$
\begin{equation}\label{eq:Normal}
	\begin{split}
		C_{\mathtt{Curvature}}(\mathcal{P}', \mathcal{P}) = \frac{1}{n'} \sum_{\vp' \in \mathcal{P}'}  \left\| \kappa'(\vp'; \mathcal{P}', \mathcal{P}) - \kappa(\vp; \mathcal{P}) \right\|_2^2 \ \
		\\ \mathrm{s.t.} \ \ \vp = \arg\min_{\vp \in \mathcal{P}} \| \vp' - \vp \|_2 ,
	\end{split}
\end{equation}
where we write $\kappa'(\vp'; \mathcal{P}', \mathcal{P})$ since the normal vector involved in its computation is from the corresponding one of $\mathcal{P}$.
Note that terms similar to (\ref{eq:disCurv}) are also used in \cite{wang2018pixel2mesh,TangCVPR2019} for single-view surface reconstruction.
Our use of the term (\ref{eq:disCurv}) in (\ref{eq:Normal}) is to encourage the consistency of local surface geometries between $\mathcal{P}'$ and $\mathcal{P}$, rather than to directly minimize (\ref{eq:disCurv}) as in \cite{wang2018pixel2mesh,TangCVPR2019}.

\vspace{0.1cm}\noindent\textbf{The Combined Geometry-Aware Objective --}
We use the following combined objective to learn to perturb individual points of $\mathcal{P}$ to obtain $\mathcal{P}'$
\begin{equation}\label{eq:Geo}
	\begin{split}
		C_{Geometry}(\mathcal{P}', &\mathcal{P})  = C_{\mathtt{Chamfer}}(\mathcal{P}', \mathcal{P}) +
		\\& \lambda_1\cdot C_{\mathtt{Hausdorff}}(\mathcal{P}', \mathcal{P}) + \lambda_2\cdot C_{\mathtt{Curvature}}(\mathcal{P}', \mathcal{P}) ,
	\end{split}
\end{equation}
where $\lambda_1$ and $\lambda_2$ are penalty parameters whose default values are set as $\lambda_1 = 0.1$ and $\lambda_2 = 1$, which work well in all our experiments; a sensitivity analysis on how their settings affect empirical performance is also presented in Section \ref{SecExps}. The combined objective $C_{Geometry}(\mathcal{P}', \mathcal{P})$ (\ref{eq:Geo}), shortened as $C_{Geo}(\mathcal{P}', \mathcal{P})$, will be used as a regularizer to penalize a misclassification loss, as specified in (\ref{eq:GeoAdvObj}).

\section{Generation of Adversarial Point Clouds}\label{SecAdvPCGen}

In the literature of adversarial 2D images, there exist generally two settings respectively termed as untargeted attack \cite{goodfellow2014explaining,moosavi2015deepfool,su2019one} and targeted attack \cite{szegedy2013intriguing,rozsa2016adversarial,papernot2015limitations}.
Assume the true label of an image signal $\bm{x}$ is $y \in \mathcal{Y}$, and $\bm{x}$ is correctly classified by $f(\cdot)$, \textit{i.e.}, $f(\bm{x}) = y$.
Untargeted attack generates an adversarial signal $\bm{x}'$ such that $f(\bm{x}') \neq y$.
Targeted attack generates $\bm{x}'$ such that $f(\bm{x}') = \hat{y}$,  with $\hat{y} \in \mathcal{Y}$ but $\hat{y} \neq y$.
It is obvious that targeted attack is a more involved task setting than the untargeted one, by crafting specified malicious signals.
In this work, we focus on the setting of targeted attack for point cloud data, which generates $\mathcal{P}'$ such that $\mathcal{P}'$ is classified as a specified class $\hat{y} \neq y$.

Technically, we adapt the state-of-the-art framework of C\&W attack \cite{CW} to achieve the goal.
Let the classification model $f(\cdot)$ be realized by a function $\bm{g}: \mathcal{X} \rightarrow \mathbb{R}^{|\mathcal{Y}|}$, and we have $f(\bm{x}) = \arg\max_{i\in\mathcal{Y}} g_i(\bm{x})$, where $g_i(\cdot)$ takes the $i^{th}$ element of $\bm{g}(\cdot)$.
When implementing $f(\cdot)$ as a deep classification network, $\bm{g}(\cdot)$ outputs the network logits, \textit{i.e.,} output of the network before the final softmax. C\&W attack commonly uses a margin based loss for the misclassification term in the general adversarial objective (\ref{eq:OverallObjective}), which gives $C_{Mis}(\bm{x}') = \max\{\max_{i\neq \hat{y}}g_i(\bm{x}') - g_{\hat{y}}(\bm{x}'), 0\}$.
This margin based $C_{Mis}(\bm{x}')$ would stop pursuing more malicious signals once $\max_{i\neq \hat{y}}g_i(\bm{x}') - g_{\hat{y}}(\bm{x}') \leq 0$.
In this work, we use the following misclassification loss to increase the malicious levels possibly achieved by targeted attack of point clouds
\begin{equation}\label{eq:GeoAdvMisClassTerm}
	C_{Mis}(\mathcal{P}') = - \log \left( \exp(g_{\hat{y}}( \mathcal{P}')) / \sum_{i=1}^{|\mathcal{Y}|} \exp(g_i( \mathcal{P}')) \right) ,
\end{equation}
where $\hat{y}$ is the specified class of targeted attack.
$C_{Mis}(\mathcal{P}')$ together with our proposed geometry-aware objective (\ref{eq:Geo}) gives our proposed method of Geometry-Aware Adversarial Attack ($GeoA^3$) for learning adversarial point cloud $\mathcal{P}'$ from $\mathcal{P}$
\begin{equation}\label{eq:GeoAdvObj}
	\min_{\mathcal{P}'} C_{GeoA^3}(\mathcal{P}', \mathcal{P}) = C_{Mis}(\mathcal{P}') +  \beta\cdot C_{Geo}(\mathcal{P}', \mathcal{P}) ,
\end{equation}
where $\beta$ is a penalty parameter controlling the overall level of geometry-aware regularization, whose setting follows \cite{CW} and is automatically adjusted via binary search.
Minimization of (\ref{eq:GeoAdvObj}) can be simply achieved by Stochastic Gradient Descent (SGD) (or its variants \cite{kingma2014adam}), which gives the following rule to update $\mathcal{P}_{t+1}'$ from $\mathcal{P}_t'$
\begin{equation}\label{eq:SGDIterUptRule}
\mathcal{P}_{t+1}' \leftarrow \mathcal{P}_t' - \eta \cdot \nabla C_{GeoA^3}(\mathcal{P}_t', \mathcal{P})  ,
\end{equation}
where $\eta$ is the learning rate. We note that without our proposed (\ref{eq:Geo}), the more aggressive misclassification loss (\ref{eq:GeoAdvMisClassTerm}) would produce $\mathcal{P}'$ whose modification from $\mathcal{P}$ could be clearly perceived.
It is our combined use of (\ref{eq:GeoAdvMisClassTerm}) and geometry-aware regularizer (\ref{eq:Geo}) that enables pursuing more adversarial, arguably less defendable, point clouds without introducing noticeable modifications.
In practice, we generate an adversarial $\mathcal{P}'$ by minimizing (\ref{eq:GeoAdvObj}) to perturb individual points $\{ \bm{p}_i \}_{i=1}^n$ contained in $\mathcal{P}$, which technically means an iterative optimization of adversarial point-wise coordinate offsets. We conduct thorough empirical studies in Section \ref{SecExps} to verify our proposed $GeoA^3$ objective.

\section{Towards Generation of Adversarial Surface Shapes}\label{SecPhysicalAttack}

An input $\mathcal{P}$ is usually obtained by sampling a discrete set of points from a certain object surface.
Consequently, adversarial effects achieved by point perturbation may be attributed to coupled factors of \emph{surface shape deformation} and \emph{point cloud re-sampling}, since the resulting $\mathcal{P}'$ may represent a discrete sampling from a deformed surface shape, a re-sampling from the same original shape, or a mixture of them. While our main focus of this paper is to generate from a given $\mathcal{P}$ an adversarial $\mathcal{P}'$ with respect to a model $f(\cdot)$, which would practically cause safety-critical issues by using $\mathcal{P}'$ to guide a LiDAR spoofer \cite{LiDARSpoofer} to fool the 3D sensor, it is more desirable to make the underlying surface $\mathcal{S}'$ represented by $\mathcal{P}'$ adversarial to the surface $\mathcal{S}$ represented by $\mathcal{P}$.
Indeed, when $\mathcal{P}'$ only represents an adversarial sampling of the original $\mathcal{S}$, it would be easily defended by a benign re-sampling.
Technically, this desirable objective is to obtain a $\mathcal{P}'$ such that when re-sampling a $\mathcal{Q}'$ from a surface reconstructed by $\mathcal{P}'$ (\emph{e.g.}, via meshing from $\mathcal{P}'$), $\mathcal{Q}'$ still has a certain degree of adversarial effect to attack the model $f(\cdot)$.

To this end, we propose a method that both explicitly promotes adversarial deformation of the surface shape, and prevents the adversarial result to be a re-sampling of the original shape.
Specifically, our design for prevention of an adversarial re-sampling from the original surface is motivated by a criterion of \emph{uniform surface point distribution} used in a state-of-the-art point set upsampling method of PU-GAN \cite{li2019pu}; the criterion promotes a uniform distribution of points in an upsampled point set.
In our context of generating an adversarial point cloud, an uniformity-promoting objective would prevent the resulting point cloud to achieve its adversarial effect simply by point perturbations that \emph{irregularly} re-distribute the points on the surface.
Intuitively speaking, the criterion uses an averaged \emph{local density of points} over a large surface area (e.g., as large as the whole surface) to guide the learning of point perturbations, such that after perturbations, the density on each local surface area is close to the averaged one. Assume that all the points of an adversarial $\mathcal{P}' = \{ \bm{p}_i' \}_{i=1}^n$ fall in a most compact, enclosing ball of radius $R$. For any $\bm{p}_i' \in \mathcal{P}'$, the criterion uses a small ball of radius $r$, centered at $\bm{p}_i'$, to include a number $m_i$ of nearby points. Let the average number of points in such balls as $\bar{m} = n r^2 / R^2$, we have the following objective to promote uniform distribution of points
\begin{equation}\label{eq:Uniform}
\begin{split}
C_{\mathtt{Uniformity}}(\mathcal{P}')  = \frac{1}{n} \sum_{i=1}^n \frac{ (m_i - \bar{m})^2 }{ \bar{m} } \cdot \frac{1}{m_i} \sum_{j=1}^{m_i} \frac{ (d_{i, j}^{\textrm{min}} - \bar{d}_i)^2 }{ \bar{d}_i } ,
\end{split}
\end{equation}
where $d_{i, j}^{\textrm{min}}$ represents the distance between any $j^{th}$ point in the ball centered at $\bm{p}_i'$ and its closest point in the ball, and we compute $\bar{d}_i = \sqrt{ 2\pi r^2 / \sqrt{3} m_i  }$; thus (\ref{eq:Uniform}) promotes an even distribution of points in each local ball. In practice, PU-GAN improves the efficiency of (\ref{eq:Uniform}) by sampling a subset of points from all the $n$ ones in $\mathcal{P}'$, and it also tunes the radius $R$ to include a large area of the surface, instead of the whole one. We also follow the practice in this work.
We augment our geometry-aware objective (\ref{eq:Geo}) with the new term (\ref{eq:Uniform}), resulting in
\begin{equation}\label{eq:GeoP}
    \begin{split}
    C_{Geo+}(\mathcal{P}', \mathcal{P})  = C_{Geo}(\mathcal{P}', \mathcal{P}) + \lambda_3 \cdot C_{\mathtt{Uniformity}}(\mathcal{P}') ,
    \end{split}
\end{equation}
where we set the penalty parameter as the default value of $\lambda_3 = 1.0$, which works well in all our experiments.
Combining (\ref{eq:GeoP}) with the misclassification loss $C_{Mis}(\mathcal{P}')$ gives our new objective of geometry-aware adversarial attack towards generation of adversarial surface shapes
\begin{equation}\label{eq:GeoAdvObjP}
	\min_{\mathcal{P}'} C_{Geo_{+}A^3}(\mathcal{P}', \mathcal{P}) = C_{Mis}(\mathcal{P}') +  \beta\cdot C_{Geo+}(\mathcal{P}', \mathcal{P}) .
\end{equation}
To optimize the objective (\ref{eq:GeoAdvObjP}), we propose an algorithm termed Iterative Normal Projection (IterNormPro), which aims for an explicit adversarial deformation of the surface shape.
Let $\mathcal{P}_t'$ denote the intermediate result of the adversarial point cloud $\mathcal{P}'$ at iteration $t$.
We first update $\mathcal{P}_t'$ via SGD as
\begin{eqnarray}\label{eq:advSurfaceIntermediateUpt}
    \widetilde{\mathcal{P}}_{t+1}' \leftarrow \mathcal{P}_t' - \eta \cdot \nabla C_{Geo_{+}A^3}(\mathcal{P}_t', \mathcal{P}) ,
\end{eqnarray}
where $\eta$ is the learning rate.
As discussed above, the updated $\widetilde{\mathcal{P}}_{t+1}'$ may account for a mixture of \emph{deformation} and \emph{point re-sampling} of the underlying surface.
To make the adversarial update account more for surface deformation, we do the following.
For any $\widetilde{\bm{p}}_{t+1}' \in \widetilde{\mathcal{P}}_{t+1}'$, we find its closest point in the input $\mathcal{P}$ by $\vp(\widetilde{\bm{p}}_{t+1}') = \arg\min_{\vp \in \mathcal{P}} \| \widetilde{\bm{p}}_{t+1}' - \vp \|_2 $, and compute the iteratively accumulated perturbation $\Delta\widetilde{\bm{p}}_{t+1}' = \widetilde{\bm{p}}_{t+1}' - \vp(\widetilde{\bm{p}}_{t+1}')$.
Assume that the normal vector $\bm{n}_{\bm{p}}$ associated with any $\bm{p} \in \mathcal{P}$ has been pre-computed (Section \ref{SecGeoAwareObj} describes the way of computing normal vectors).
By retrieving the normal vector $\vn_{\vp}(\widetilde{\bm{p}}_{t+1}')$ associated with the nearest $\vp(\widetilde{\bm{p}}_{t+1}')$, we are able to project the perturbation $\Delta\widetilde{\bm{p}}_{t+1}'$ onto the normal direction that is ideally \emph{locally orthogonal} to the underlying surface, giving rise to the adversarial update at iteration $t+1$ as
\begin{eqnarray}\label{eq:advSurfaceProjection}
    \vp_{t+1}' \leftarrow \vp(\widetilde{\bm{p}}_{t+1}') + \left<\Delta\widetilde{\bm{p}}_{t+1}', \vn_{\vp}(\widetilde{\bm{p}}_{t+1}') \right> \cdot \frac{\vn_{\vp}(\widetilde{\bm{p}}_{t+1}')}{\|\vn_{\vp}(\widetilde{\bm{p}}_{t+1}')\|_2} .
\end{eqnarray}
We apply the above procedure to all $\{ \bm{p}_t' \in \mathcal{P}_t' \}$, and have the updated adversarial point cloud $\mathcal{P}_{t+1}' = \{ \bm{p}_{t+1}' \}$.
The final result is obtained as $\mathcal{P}' = \mathcal{P}_T'$, given a total of $T$ iterations.
We term the new method that optimizes the objective (\ref{eq:GeoAdvObjP}) via iterative normal projection as $Geo_{+}A^3$-IterNormPro.
The method can be intuitively thought as deforming the underlying surface in a local contraction or expansion manner.
Given the geometry-aware objective (\ref{eq:GeoP}) that promotes smooth and uniform distribution of surface points, $Geo_{+}A^3$-IterNormPro is expected to produce less perceptible adversarial results whose adversarial effects come more from shape deformations.

\vspace{0.1cm}\noindent\textbf{Practical Implementation and Physical Attack ---}
Practically, we employ Points2Surf reconstruction \cite{erler2020points2surf} to reconstruct a surface mesh $\mathcal{S}'$ from a given $\mathcal{P}'$, and uniformly sampling points $\mathcal{Q}'$ on the surface of the mesh $\mathcal{S}'$ with probability proportional to the face area as that in {\cite{PointNetPP}}.
For physical attack, we do 3D printing using $\mathcal{S}'$ and then scan the printed object to obtain $\mathcal{Q}'$.
The respectively obtained $\mathcal{Q}'$ is used for evaluation of attack performance.

\section{Experiments}
\label{SecExps}

\begin{table*}[htbp]
    \caption{ Ablation studies on our proposed $GeoA^3$ (\ref{eq:GeoAdvObj}).
        Each input, benign point cloud contains $1,024$ points.
        Performance is measured in terms of both the attack success rate (\%) and geometric regularity $R(\mathcal{P}')$ (\ref{Eq:GeoRegularityMeasure}).
        Attack success rates are reported by dropping a range of ratios of points from their adversarial results using the state-of-the-art SOR defense method \cite{Deflecting}.
        Experiments reported here are conduced using PointNet \cite{PointNet} as the model of classifier.
        More results on PointNet++ \cite{PointNetPP} and DGCNN \cite{DGCNN} are of similar comparative quality; they are included in the Appendix.
	}\label{table:ablation_study_per}
	\begin{center}
		\begin{small}
			\scalebox{0.93}{
			\begin{tabular}{l c c c c c c c c}
				\toprule
				\multirow{2}{*}{Method}	&
				\multicolumn{7}{c}{\makecell{Attack success rate (\%) \\ defense by dropping different ratios of points via SOR \cite{Deflecting}}}
				& \multirow{2}{*}{ \makecell{Geometric regularity \\ $R(\mathcal{P}')$}}\\
				\cline{2-8}
				~ & $0\%$ & $1\%$ & $2\%$ & $5\%$ & $10\%$ & $15\%$ & $20\%$ & ~ \\
				\midrule
				$GeoA^3$ with all three geometry-aware terms &  $100$ & $\bm{83.47}$ & $\bm{70.56}$ & $\bm{52.61}$ & $\bm{31.58}$ & $\bm{18.62}$ & $\bm{11.71}$ &$\bm{0.0968}$ \\
				\midrule
				{w/o Chamfer Distance} & $100$ & $82.17$ & $68.14$ & $50.08$ & $30.24$ & $15.14$ & $10.01$ & $0.1020$ \\
				{w/o Hausdorff Distance} & $100$ & $24.51$ & $18.89$ & $11.68$ & $6.60$ & $3.97$ & $3.12$ & $0.1588$\\
				{w/o Consistency of Local Curvatures}  &  $100$ & $53.85$ & $35.90$ & $14.88$ & $5.57$ & $2.81$ & $2.00$ & $0.1055$ \\
				\bottomrule
			\end{tabular}
			}
		\end{small}
	\end{center}
\end{table*}

\noindent\textbf{Dataset ---}
We use object instances from ModelNet40 \cite{wu20153d} to evaluate our proposed methods.
The dataset consists of $12,311$ CAD models belonging to $40$ semantic categories.
To have a point cloud, we follow \cite{PointNetPP} and sample points uniformly from each CAD model and then normalize them into a unit ball.
We use the official data splits of ModelNet40 for point set classification \cite{PointNet,PointNetPP}, which give $9,843$ instances to train classifiers that are to be attacked.
For adversarial point cloud attack, we follow \cite{CW3DAdv} and randomly select $25$ instances for each of $10$ object categories in the ModelNet40 testing set, which can be well classified by the classifiers of interest.
The 10 object categories include \textit{airplane, bed, bookshelf, bottle, chair, monitor, sofa, table, toilet,} and \textit{vase}.
For surface-level attack, we take all the instances of the 40 categories that are well classified in the ModelNet40 testing set.

\begin{figure*}[htbp]
	\centering
	\begin{minipage}{0.99\textwidth}
		\centering
		\includegraphics[width=0.99\textwidth]{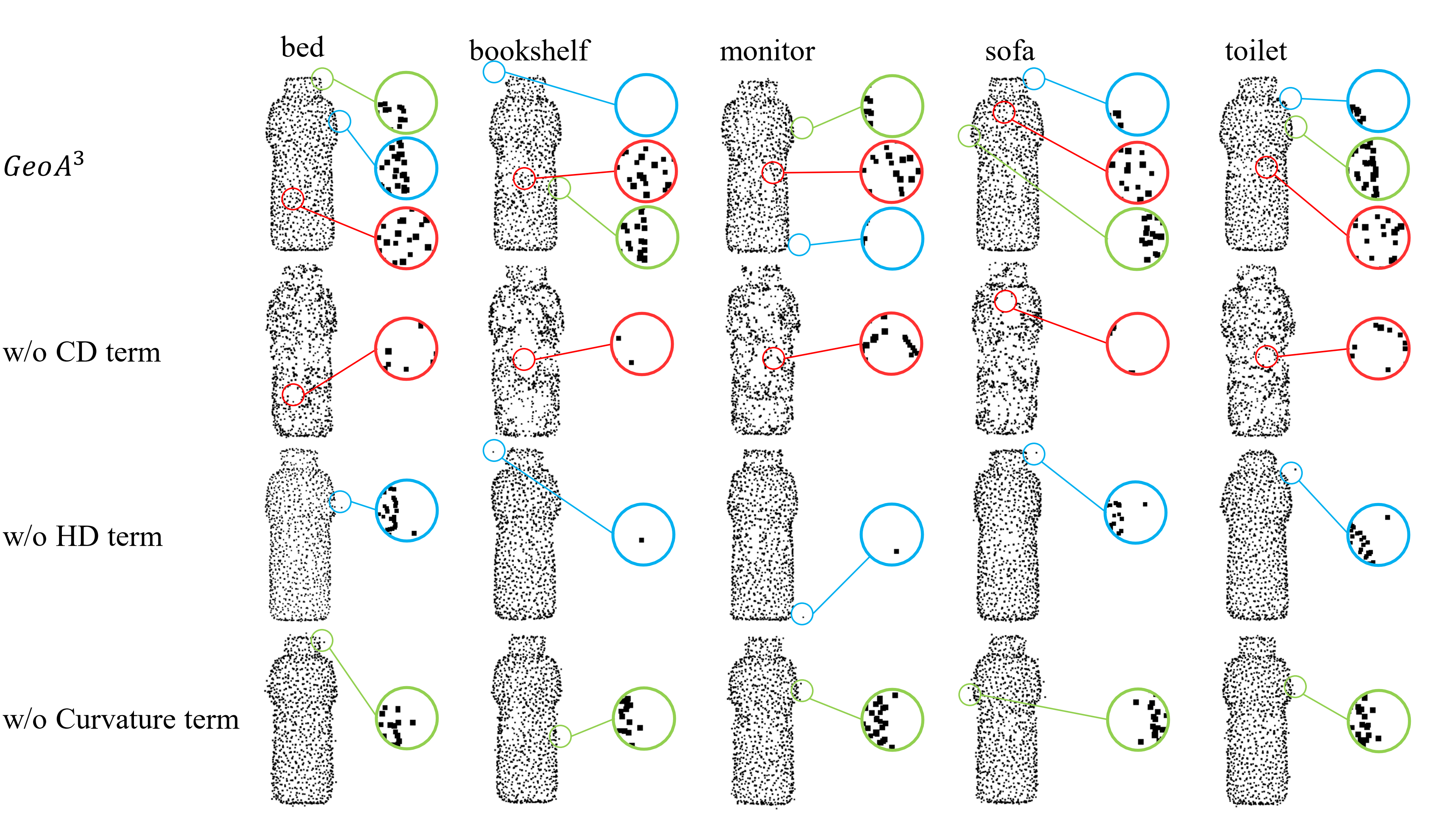}
	\end{minipage}
	\caption{
        Results of ablation study by removing individual terms from our proposed geometry-aware regularizer (\ref{eq:Geo}), where CD stands for the term of Chamfer distance, HD for that of Hausdorff distance, and Curvature for that of consistency of local curvatures.
        Each input, benign point cloud contains $1,024$ points, to which adversarial attacks are applied.
        Results are from a \emph{bottle} instance that is attacked against PointNet \cite{PointNet}, targeting at the categories of \emph{bed}, \emph{bookshelf}, \emph{monitor}, \emph{sofa}, and \emph{toilet}.
        All the shown examples conduct the attacks successfully.
        Lens of different colors are used to highlight the differences among the comparative methods; those of the same color refer to the same surface area located on different results.
        Comparative results from instances of other categories are of similar quality.}
\label{fig:ablation_study_nonpart}
\end{figure*}

\begin{table*}[htbp]
    \caption{
        Ablation studies on the misclassification loss (\ref{eq:GeoAdvMisClassTerm}) used in our proposed $GeoA^3$.
        Performance is measured in terms of both the attack success rate (\%) and geometric regularity $R(\mathcal{P}')$.
        Attack success rates are reported by dropping a range of ratios of points from their adversarial results using the state-of-the-art defense method of SOR \cite{Deflecting}.
        All experiments are conduced using PointNet \cite{PointNet} as the model of classifier
    }
    \label{table:ablation_miscls}
    \begin{center}
        \begin{small}
        \scalebox{0.90}{
        \begin{tabular}{l c c c c c c c c}
            \hline
            \multirow{2}{*}{Method} & \multicolumn{7}{c}{\makecell{Attack success rate($\%$) \\ defense by dropping different ratios of points via SOR}} & \multirow{2}{*}{ \makecell{Geometric regularity \\ $R(\mathcal{P}')$}} \\
            \cline{2-8}
            ~ & $0\%$ & $1\%$ & $2\%$ & $5\%$ & $10\%$ & $15\%$ & $20\%$ & ~ \\
            \hline
            $GeoA^3$ & $100$ & $\bm{83.47}$ & $\bm{70.56}$ & $\bm{52.61}$ & $\bm{31.58}$ & $\bm{18.62}$ & $\bm{11.71}$ &$0.0968$ \\
            \hline
            $GeoA^3$ replaced with margin-based loss & $100$ & $78.62$ & $67.42$ & $44.98$ & $21.16$ & $9.29$ & $4.44$ & $\bm{0.0819}$ \\
            Xiang \emph{et al.} \cite{CW3DAdv} replaced with aggressive loss & $100$ & $3.21$ & $1.60$ & $1.02$ & $0.62$ & $0.36$ & $0.31$ & $0.1956$\\
            \hline
        \end{tabular}
        }
        \end{small}
    \end{center}
\end{table*}

\begin{table*}[htbp]
	\caption{Adversarial results when respectively sampling different numbers of points from each CAD model as the working point clouds. Performance is measured in terms of both the attack success rate (\%) and geometric regularity $R(\mathcal{P}')$ (\ref{Eq:GeoRegularityMeasure}). Attack success rates are reported by dropping a range of ratios of points from their adversarial results using the state-of-the-art SOR defense method \cite{Deflecting}. All experiments are conduced using PointNet \cite{PointNet} as the model of classifier.}
\label{table:ablation_study_num_points}
	\begin{center}
		\begin{small}
			\scalebox{1.00}{
			\begin{tabular}{l c c c c c c c c}
				\toprule
				\multirow{2}{*}{Number of points} & \multicolumn{7}{c}{\makecell{Attack success rate (\%) \\ defense by dropping different ratios of points via SOR \cite{Deflecting}}}
				& \multirow{2}{*}{ \makecell{Geometric regularity \\ $R(\mathcal{P}')$}}\\
				\cline{2-8}
				~ & $0\%$ & $1\%$ & $2\%$ & $5\%$ & $10\%$ & $15\%$ & $20\%$ & ~ \\
				\midrule
				$1,024$ &  $100$ & $83.47$ & $70.56$ & $52.61$ & $31.58$ & $18.62$ & $11.71$ &$0.0968$ \\
				$2,048$ &  $100$ & $81.02$ & $70.44$ & $48.36$ & $28.04$ & $16.58$ & $10.62$ & $0.0988$ \\
				$4,096$ &  $100$ & $80.80$ & $71.11$ & $47.73$ & $22.31$ & $18.04$ & $9.87$ & $0.0997$ \\
				\bottomrule
			\end{tabular}
			}
		\end{small}
	\end{center}
\end{table*}

\noindent\textbf{Evaluation Metrics ---}
To quantitatively compare the adversarial results generated by different methods, we use both the measure of attack success rate and that of geometric regularity.
The former measures, among benign point clouds, the ratio of their adversarial results that successfully fool a classifier; note that all the benign point clouds to be attacked are classified correctly by the classifier.
We report attack success rates under a state-of-the-art defense method of Statistical Outlier Removal (SOR) \cite{Deflecting}, which works by statistically dropping certain points from any adversarial $\mathcal{P}'$.
For the latter measure, we are inspired by the desired surface property of \emph{fairness} \cite{botsch2010polygon}, and introduce the following discrete and approximate measure to quantify the regularity of a point cloud $\mathcal{P}$
\begin{equation}\label{Eq:GeoRegularityMeasure}
	R(\mathcal{P}) = \max_{\vp \in \mathcal{P} } \frac{1}{k} \sum_{\vq\in \mathcal{N}_{\vp}} \| \vq -  \textrm{proj}_{\mathcal{T}(\mathcal{N}_{\vp})}\vq \|_2 ,
\end{equation}
which computes the $l_2$-norm distance between any neighboring point $\vq \in \mathcal{N}_{\vp}$ and its projection onto the tangent plane $\mathcal{T}(\mathcal{N}_{\vp})$ computed from the neighborhood $\mathcal{N}_{\vp}$.
Note that a surface is generally considered fair if its curvatures are globally minimized.
The proposed measure (\ref{Eq:GeoRegularityMeasure}) is easily computable, and functions as an approximate surrogate of the globally maximum curvature defined on the discrete point cloud, since a surface with high values of local curvature gives a high value of $R(\mathcal{P})$.
The proposed $R(\mathcal{P})$ is also relevant to the measure of perceptual aesthetics \cite{MiuraAesthetics,LigangAesthetics}, since both of them are motivated from the same desired surface property of fairness.
A better-performing method is expected to generate adversarial results that achieve higher attack success rates and lower values of $R(\mathcal{P}')$, simultaneously.
We also conduct subjective studies for a qualitative evaluation by collecting human preferences from Amazon Mechanical Turk.

\noindent\textbf{Models and Implementation Details ---}
We use PointNet \cite{PointNet}, PointNet++ \cite{PointNetPP}, and DGCNN \cite{DGCNN} as our classifiers of interest, against which comparative methods fire attacks.
Without mentioning otherwise, all the methods follow a white-box, targeted attack protocol via point perturbation.
We use Adam \cite{kingma2014adam} to optimize the objective (\ref{eq:GeoAdvObj}) of our proposed $GeoA^3$; we use a fixed learning schedule of $500$ iterations, where the learning rate and momentum are respectively set as $0.01$ and $0.9$.
We set $k = 16$ to define local point neighborhoods.
We use the default values of $\lambda_1 = 0.1$ and $\lambda_2 = 1.0$ for penalty parameters in (\ref{eq:Geo}), the default one of $\lambda_3 = 1.0$ for that in (\ref{eq:GeoP}).
The penalty $\beta$ in (\ref{eq:GeoAdvObj}) is initialized as $2,500$ and automatically adjusted by conducting 10 runs of binary search, which follows \cite{CW}.


\subsection{Ablation Studies and Sensitivity Analysis}\label{SecExpsAblation}

\vspace{0.1cm}
\noindent\textbf{Ablation Studies ---}
To test the efficacy of our proposed $GeoA^3$ (\ref{eq:GeoAdvObj}), we first conduct ablation studies by removing individual terms in the geometry-aware regularizer (\ref{eq:Geo}) and compare the results quantitatively and qualitatively. Table \ref{table:ablation_study_per} shows that across a range of dropping ratios via the state-of-the-art defense SOR \cite{Deflecting}, attack success rates drop by removing any one of the three terms in (\ref{eq:Geo}), with the term of Hausdorff distance playing the most important role and that for consistency of local curvatures following. The observations are consistent for the measure of geometric regularity. Examples of qualitative results in Figure \ref{fig:ablation_study_nonpart} also show that removing any of the three terms will produce adversarial results containing either less regular surfaces or clearly visible outliers. Results from instances of other categories are of similar quality. To understand why these happen, we note that without imposing proper constraints on the point perturbations, adversarial attack by minimizing the misclassification term in the $GeoA^3$ objective (\ref{eq:GeoAdvObj}) tends to take effect by overly perturbing a small portion of points in the input point cloud, instead of perturbing all the points evenly; this easily produces point outliers, as shown in Figure \ref{fig:ablation_study_nonpart}.
In other words, a small number of point outliers are the main factor that makes adversarial attacks in the case of less or no constraints.
Imposing the constraint via the term (\ref{eq:HD}) of Hausdorff distance directly penalizes the large deviation of points from the surface of any adversarial point cloud, thus suppressing the generation of point outliers.
On the other hand, given that the input, benign point cloud represents a smooth surface, the term (\ref{eq:Normal}) for consistency of local curvatures can achieve the same effect of penalizing point outliers, but it achieves it in a softer and globally more even manner.
Consequently, without using either the term of Hausdorff distance or that for consistency of local curvatures, the attack success rates drop significantly when removing points from the adversarial results via SOR, as observed in Table \ref{table:ablation_study_per}.
These comparisons confirm the combined advantage of our proposed geometry-aware regularizer (\ref{eq:Geo}) to achieve adversarial point clouds that are both less defendable and less perceptible, simultaneously.

\begin{figure}[htbp]
    \begin{center}
        \begin{minipage}{0.24\textwidth}
            \centering
            {
                \includegraphics[width=0.90\textwidth]{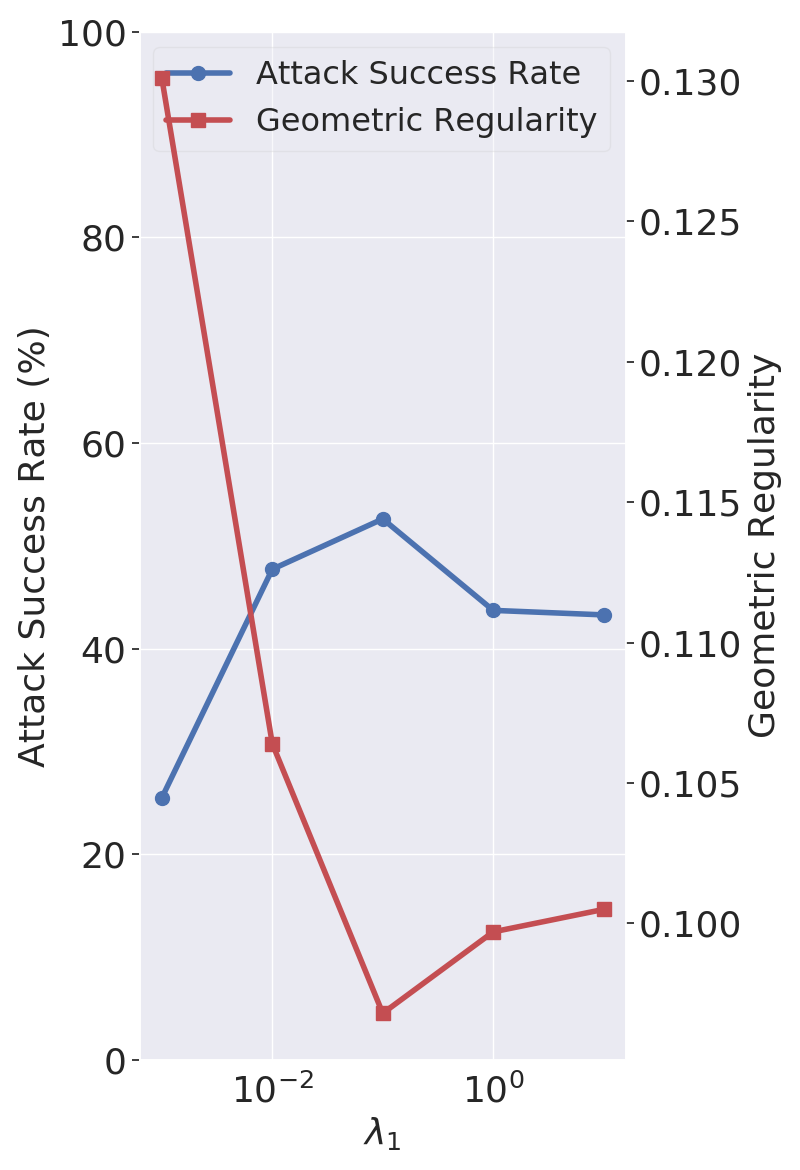}
            }
        \end{minipage}
        \begin{minipage}{0.24\textwidth}
            \centering
            {
                \includegraphics[width=0.90\textwidth]{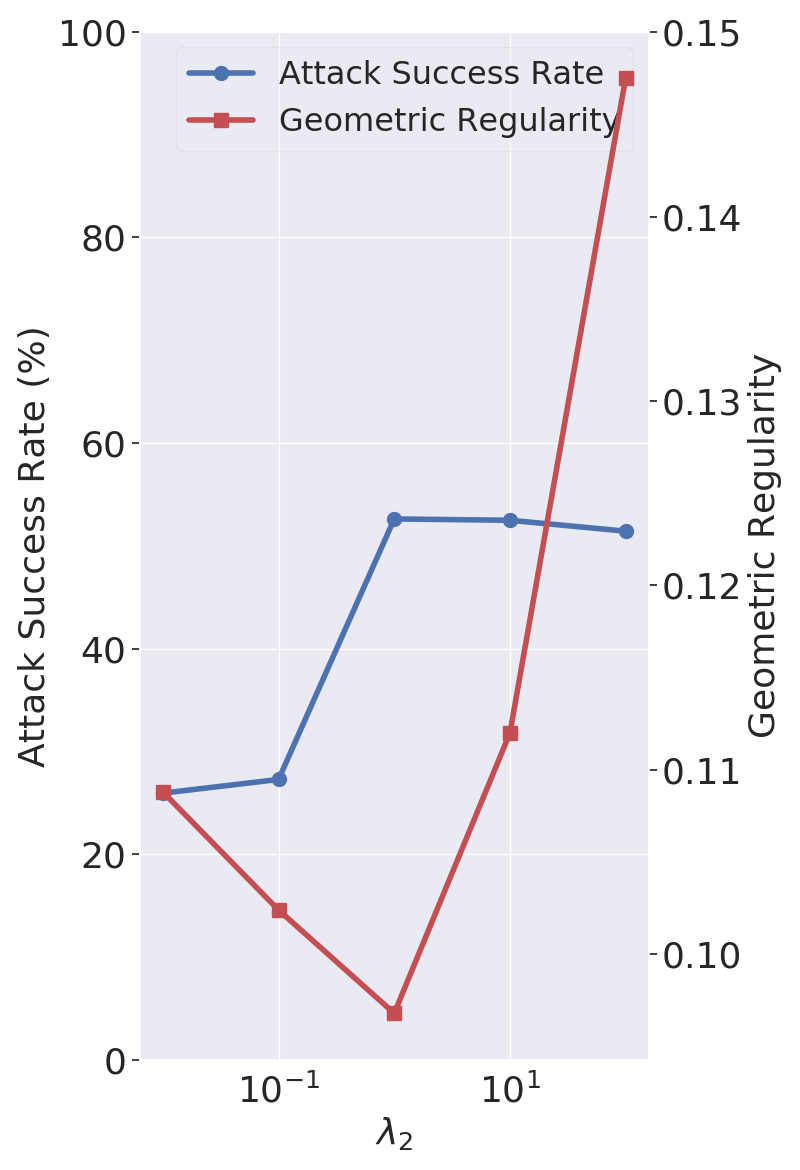}
            }
        \end{minipage}
	\end{center}
    \caption{
        Sensitivity analysis of our method with respect to the values of $\lambda_1$ and $\lambda_2$ in the geometry-aware regularizer (\ref{eq:Geo}).
        We plot both the attack success rate (\%), under the SOR defense \cite{Deflecting} of $5\%$ point dropping, and geometric regularity (\ref{Eq:GeoRegularityMeasure}) by varying the values of either $\lambda_1$ (left) or $\lambda_2$ (right) around their respective default values of $\lambda_1 = 0.1$ and $\lambda_2 = 1$.
        Experiments are conducted on point clouds of $1,024$ points using PointNet \cite{PointNet} as the model of classifier.
        }
\label{fig:sensitivity_study}
\end{figure}

Table \ref{table:ablation_study_per} also shows that Hausdorff distance plays a more important role when dropping fewer points via SOR defense.
This is reasonable since these fewer points are the outliers that mainly cause the adversarial attacks, and the term of Hausdorff distance directly suppresses their generation.
When dropping more points by SOR, however, the term for consistently of local curvatures is getting sightly more effective, since it can make the adversarial point perturbations softer and globally more evenly distributed; consequently, defense via SOR becomes more difficult.

\begin{figure}[htbp]
    \centering
    \begin{minipage}{0.32\textheight}
        \centering
        {
            \includegraphics[width=0.99\textwidth]{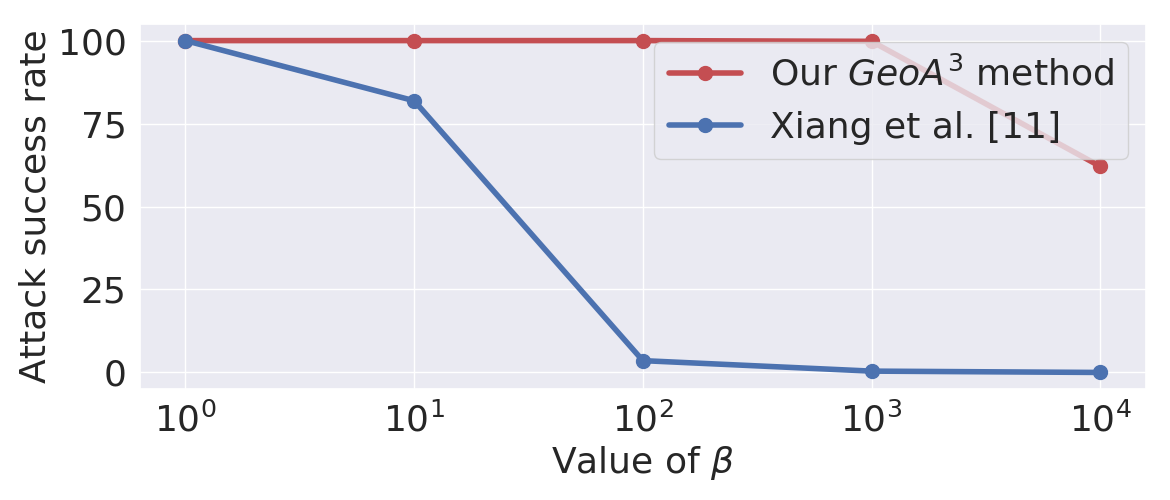}
        }
    \end{minipage}\\    
    \caption{
        Sensitivity analysis on a range of fixed $\beta$ values in the $GeoA^3$ objective (\ref{eq:GeoAdvObj}).
        Experiments are conducted on clouds of $1,024$ points using PointNet \cite{PointNet}.
        More results on PointNet++ \cite{PointNetPP} and DGCNN \cite{DGCNN} are of similar comparative quality; they are included in the Appendix.
        }
    \label{fig:beta_optimal}
\end{figure}

\begin{table*}[htbp]
	\caption{Comparative results of different methods for adversarial point clouds.
		Each input, benign point cloud contains $1,024$ points.
		Performance is measured in terms of both the attack success rate (\%) and geometric regularity $R(\mathcal{P}')$ (\ref{Eq:GeoRegularityMeasure}).
        For different methods, attack success rates are reported by dropping a range of ratios of points from their adversarial results using the state-of-the-art SOR defense method \cite{Deflecting}.
        Experiments reported here are conduced using PointNet \cite{PointNet} as the model of classifier.
        More results on PointNet++ \cite{PointNetPP} and DGCNN \cite{DGCNN} are of similar comparative quality; they are included in the Appendix.
		}\label{table:misleading_per_compare}
	\begin{center}
		\begin{small}
			\scalebox{1.0}{
				\begin{tabular}{l c c c c c c c c}
					\toprule
					\multirow{2}{*}{Method}	&
					\multicolumn{7}{c}{\makecell{ Attack success rate (\%) \\ defense by dropping different ratios of points via SOR \cite{Deflecting}}}
					& \multirow{2}{*}{ \makecell{Geometric regularity \\ $R(\mathcal{P}')$}}\\
					\cline{2-8}
					~ & $0\%$ & $1\%$ & $2\%$ & $5\%$ & $10\%$ & $15\%$ & $20\%$ & ~ \\
					\midrule
					Xiang \emph{et al.} \cite{CW3DAdv} & $100$ & $1.42$ & $1.24$ & $0.98$ & $0.80$ & $0.67$ & $0.49$ & $0.1772$\\

					Liu \emph{et al.} \cite{Extending} & $100$  & $0.63$ & $0.40$ & $0.36$ & $0.09$ & $0.09$ & $0.04$  & $0.1658$\\

					Tsai \emph{et al.} \cite{AAAICloseRelatedWork} & $100$ & $23.84$ & $17.74$ & $14.50$ & $11.12$ & $9.01$ & $8.12$ & $0.1602$\\

					$GeoA^3$ & $100$ & $\bm{83.47}$ & $\bm{70.56}$ & $\bm{52.61}$ & $\bm{31.58}$ & $\bm{18.62}$ & $\bm{11.71}$ & $\bm{0.0968}$\\
					\bottomrule
				\end{tabular}}
			\end{small}
		\end{center}
\end{table*}

\begin{figure*}[htbp]
	\begin{center}
		\includegraphics[width=0.90\textwidth]{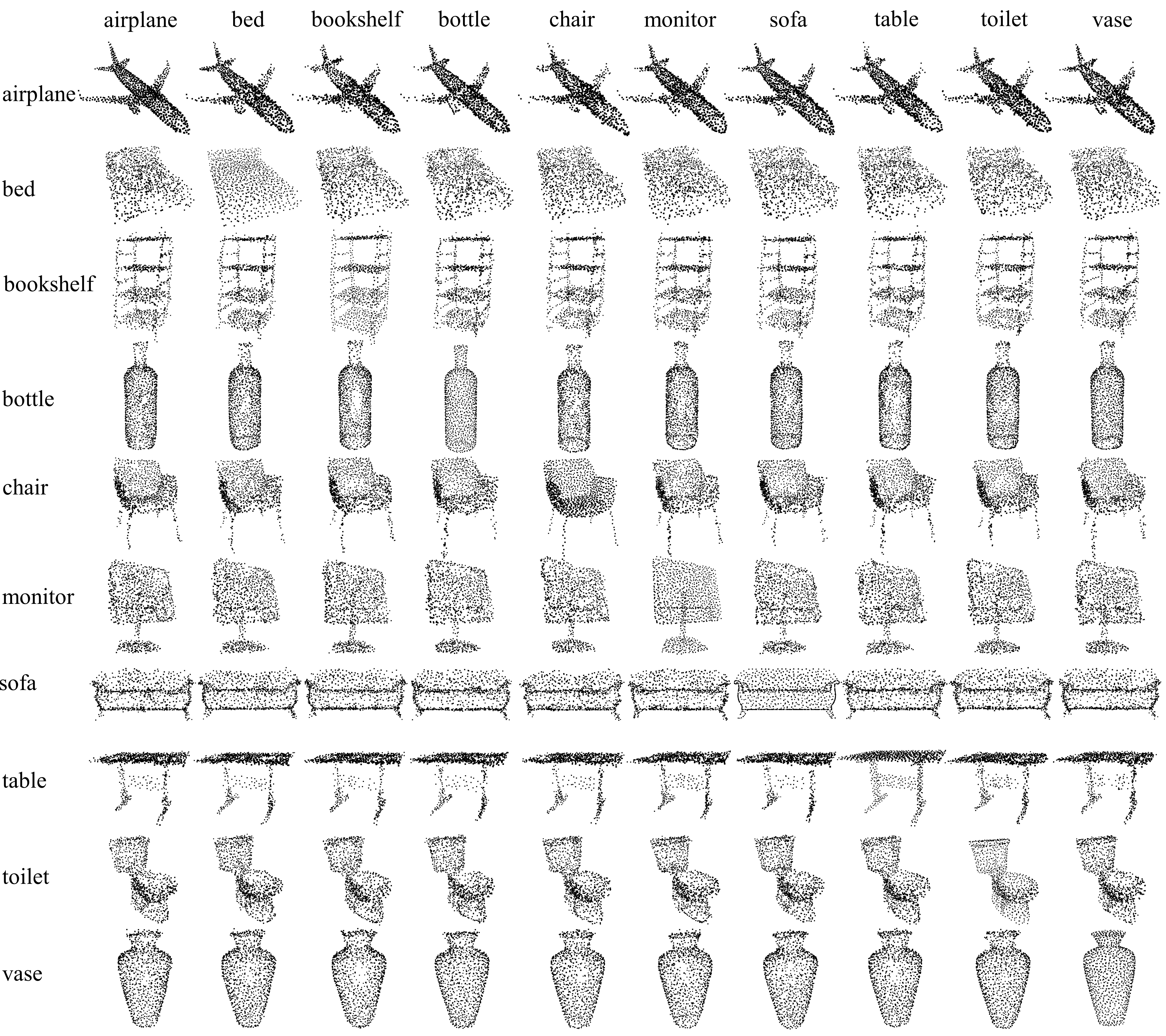}
	\end{center}
	\caption{
		Adversarial examples from our proposed $GeoA^3$.
		Results are organized in a matrix form where each off-diagonal entry presents the adversarial point cloud obtained by attacking an example instance from a certain category against PointNet \cite{PointNet}, targeting at one of the remaining 9 categories, and the diagonal entry presents the input, benign point cloud.}
 \label{fig:imper_per_tenclass}
\end{figure*}

\begin{figure*}[htbp]
	\centering
	\begin{minipage}{0.90\linewidth}
		\centering
		\includegraphics[width=0.99\textwidth]{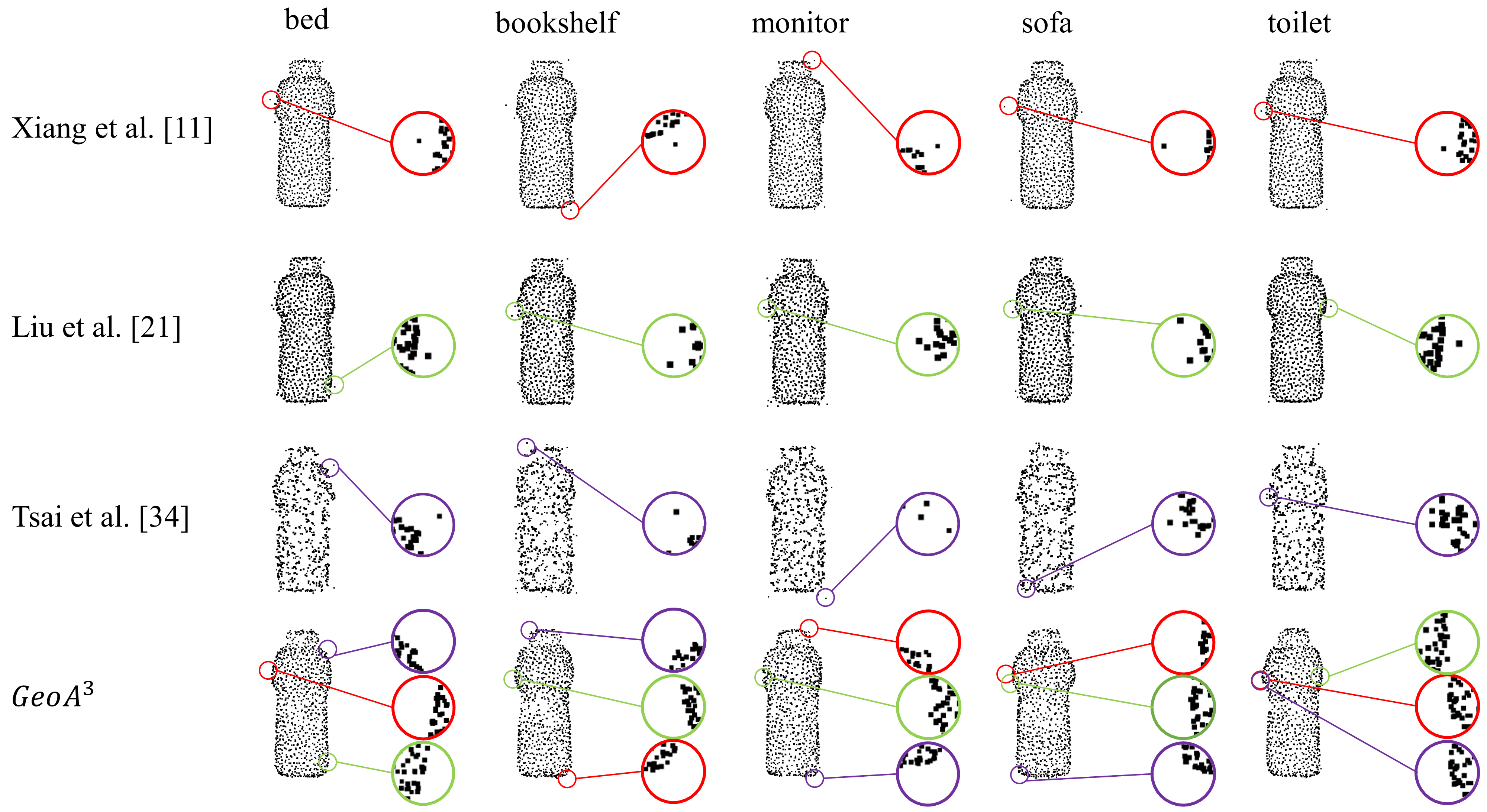}
	\end{minipage}
	\caption{
        Qualitative comparisons among adversarial results generated by different methods. Each point cloud contains $1,024$ points.
		Results are from a \emph{bottle} instance that is attacked against PointNet \cite{PointNet}, targeting at the categories of \emph{bed}, \emph{bookshelf}, \emph{monitor}, \emph{sofa}, and \emph{toilet}.
		All the shown results can successfully make the targeted attacks.
        Lens of different colors are used to highlight the differences among the comparative methods; those of the same color refer to the same surface area located on different results.
        Results of different methods from other instances and targeting categories are of similar comparative quality.
		}\label{fig:imper_per}
\end{figure*}

We further perform two aspects of ablation study to validate the efficacy of our proposed aggressive misclassification loss (\ref{eq:GeoAdvMisClassTerm}).
We replace this aggressive loss in our $GeoA^3$ objective with the margin-based loss used in C$\&$W attack (cf. Section \ref{SecAdvPCGen} for its formulation), and dub the method as ``$GeoA^3$ replaced with margin-based loss''. We also replace the same margin-based misclassification loss in a comparative Xiang \emph{et al.} \cite{CW3DAdv} with our aggressive loss (\ref{eq:GeoAdvMisClassTerm}), and dub the method as ``Xiang \emph{et al.} \cite{CW3DAdv} replaced with aggressive loss''. Results in Table \ref{table:ablation_miscls} show that without using the misclassification loss (\ref{eq:GeoAdvMisClassTerm}), our $GeoA^3$ has a large drop in terms of attack success rate, while the measure of geometric regularity slightly improves, due to the use of the less aggressive, margin-based loss. Replacing the margin-based loss in Xiang \emph{et al.} with our proposed loss does not bring benefits in terms of both the attack success rate and geometric regularity; this confirms the importance of the geometry-aware objectives in our $GeoA^3$ method.

\vspace{0.1cm}
\noindent\textbf{Sensitivity Analysis on the Number of Points ---}
While most of the experiments are conducted on point clouds of $1,024$ points, it is interesting to investigate whether the adversarial effects vary with respect to different numbers of points per instance.
We conduct such experiments by respectively sampling $1,024$, $2,048$, and $4,096$ points from each CAD model as our working point clouds.
We correspondingly set their sizes of local neighborhoods as $k = 16$, $32$, and $64$.
Table \ref{table:ablation_study_num_points} shows that under both the measures of attack success rate and geometric regularity, our results are relatively stable with respect to different numbers of points.
We choose point clouds of $1,024$ points as the default setting.

\vspace{0.1cm}
\noindent\textbf{Sensitivity Analysis on Penalty Parameters ---}
To investigate how the penalty parameters $\lambda_1$ and $\lambda_2$ in the geometry-aware regularizer (\ref{eq:Geo}) affect adversarial efficacy, we conduct experiments by varying each of them around their default values, while keeping the other one fixed.
Plottings in Figure \ref{fig:sensitivity_study} show that attack success rates are relatively stable with respect to different values of $\lambda_1$ and $\lambda_2$, and our default values of $\lambda_1 = 0.1$ and $\lambda_2 = 1$ give the optimal results in terms of geometric regularity.

We follow \cite{CW} and use binary search to find the optimal value of $\beta$ when optimizing our $GeoA^3$ objective (\ref{eq:GeoAdvObj}). It is interesting to observe how our method performs when setting $\beta$ as fixed values. Figure \ref{fig:beta_optimal} conducts such investigations, where we also plot results from Xiang \emph{et al.} as a reference, which uses a same framework of $C\&W$ attack. On different classification models, the attack success rates of our $GeoA^3$ are more stable as $\beta$ varies. This is because in Xiang \emph{et al.}, $l_2$ norm is used to constrain the perturbations of individual points; consequently, large offset per point is prohibited and the point set as a whole is less flexible to be deformed in an adversarial manner.
Instead, our proposed $GeoA^3$ uses geometry-aware objectives that constrain the deformation of point set as a whole, while imposing no constraints on perturbations of individual points; our method thus allows more flexible deformations and the deformed point sets could still be adversarial when $\beta$ increases.

\subsection{Comparative Results of Adversarial Point Clouds}\label{SecExpsCompare}

In this section, we compare our proposed $GeoA^3$ with existing methods \cite{CW3DAdv,Extending,AAAICloseRelatedWork} that generate adversarial point clouds under the setting of white-box, targeted attack.
In Xiang \emph{et al.} \cite{CW3DAdv}, adversarial point perturbation follows the framework of C\&W attack \cite{CW}, by using a margin-based misclassification loss regularized by point-wise $l_2$-norms constraining the magnitudes of point perturbation.
Liu \emph{et al.} \cite{Extending} adopts a variant of the basic iterative method in \cite{kurakin2016adversarial}, under a constraint of $l_2$-norm distance between the entire, benign point cloud and the adversarial one.
Tsai \emph{et al.} \cite{AAAICloseRelatedWork} also follows the framework of C\&W attack, whose constraint combines a global Chamfer distance and a local term that encourages the compactness of local neighborhoods in the generated adversarial point cloud.
Results of these methods are obtained either by using their released codes, when available, or by reproducing their methods; in both cases, we tune their respective hyper-parameters as the optimal ones.
All the comparative experiments are conducted using PointNet \cite{PointNet} as the model of classifier.

In Table \ref{table:misleading_per_compare}, we compare different methods under the measures of attack success rate, again under the state-of-the-art SOR defense, and geometric regularity $R(\mathcal{P}')$ (\ref{Eq:GeoRegularityMeasure}).
Under both of the two measures, our proposed $GeoA^3$ performs much better, across a range of dropping ratios via SOR, than all the existing methods do.
The comparisons confirm the combined efficacy of using our more aggressive misclassification loss (\ref{eq:GeoAdvMisClassTerm}) and geometry-aware regularizer (\ref{eq:Geo}) to simultaneously achieve the strongest adversarial attack and least visual perception of the perturbations.
The later advantage of our method is illustrated in Figure \ref{fig:imper_per} where we show adversarial results of a \emph{bottle} instance that is attacked, targeting at $5$ different categories, by different methods (the same instance and targeting categories as in Figure \ref{fig:ablation_study_nonpart}).
Clearly, $GeoA^3$ gives adversarial results whose visual differences from the benign ones can arguably be least perceived by humans; results from competing methods contain either point outliers and/or local geometric irregularities.

\begin{figure}[htbp]
	\begin{center}
		\includegraphics[width=0.28\textwidth]{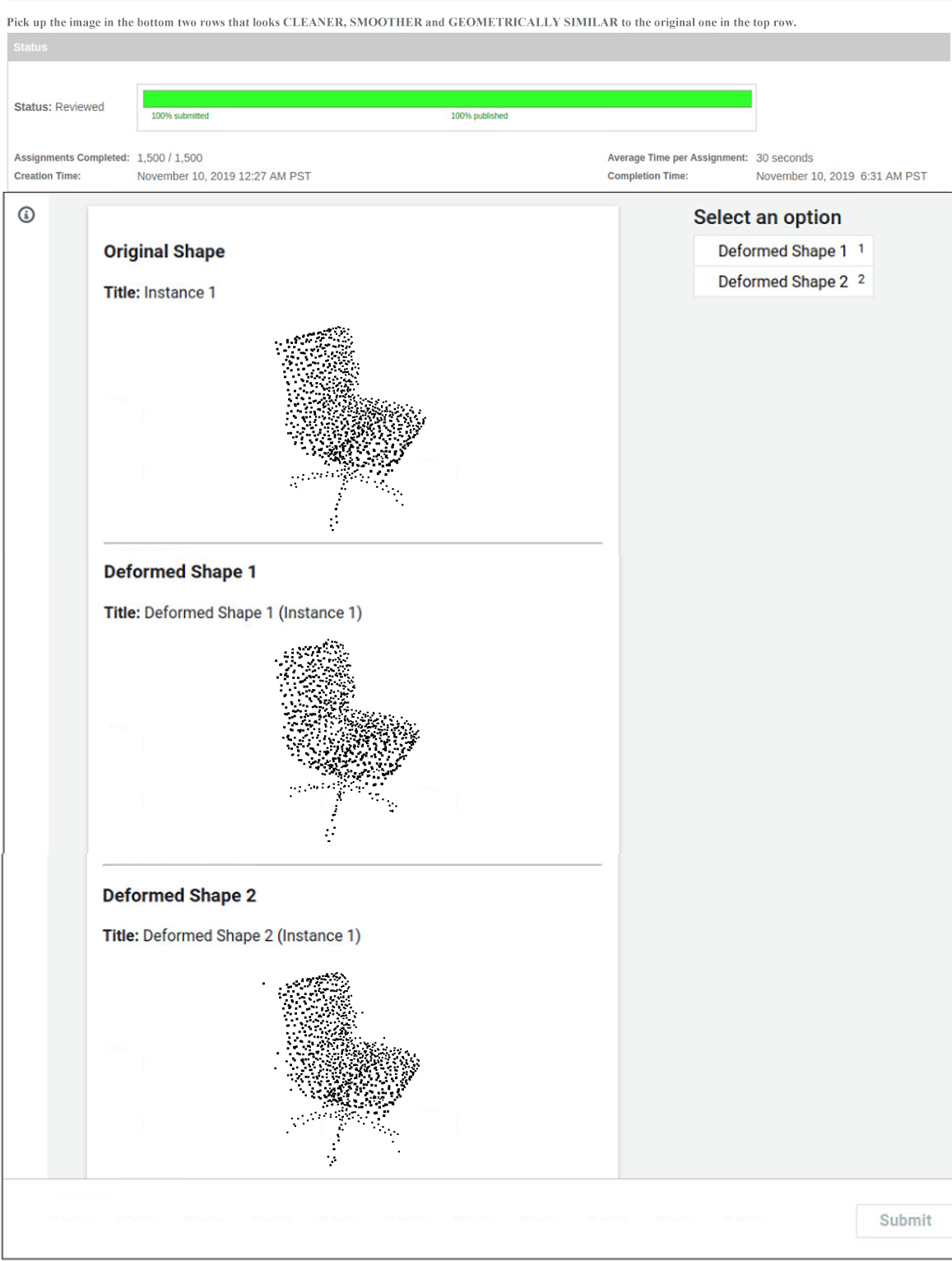}
	\end{center}
	\caption{
		The interface we use to conduct subjective user studies by uploading onto Amazon Mechanical Turk a triple of point cloud snapshots including the benign one, the adversarial one generated by Xiang \emph{et al.} \cite{CW3DAdv}, and the adversarial one generated by our $GeoA^3$.
		All the uploaded results attack PointNet \cite{PointNet} successfully.
		}
 	\label{fig:AMT_ill}
\end{figure}

\begin{table}[thbp]
    \caption{
        Results of subjective study on Amazon Mechanical Turk, by comparing our proposed $GeoA^3$ with Xiang \emph{et al.} \cite{CW3DAdv}. Given a benign point cloud of an object instance, adversarial point clouds attacking a PointNet classifier are respectively generated by the two methods. $128$ online participants are asked to compare and choose which one of the two adversarial results are visually more similar to the benign one. We allow each participant to contribute at most 30 online trials, and collect $1500$ trials in total.
    }
    \label{table:AMT}
    \begin{center}
        \begin{small}
        \scalebox{1.0}{
        \begin{tabular}{c c}
            \hline
            Method & Percentage of choice ($\%$) \\
            \hline
            Xiang \emph{et al.} \cite{CW3DAdv} & 17.94 \\
            $GeoA^3$ & $\bm{82.06}$ \\
            \hline
        \end{tabular}
        }
        \end{small}
    \end{center}
\end{table}

\begin{figure*}[htbp]
	\centering
    \begin{minipage}{0.80\textwidth}
        \centering
        \includegraphics[width=0.99\textwidth]{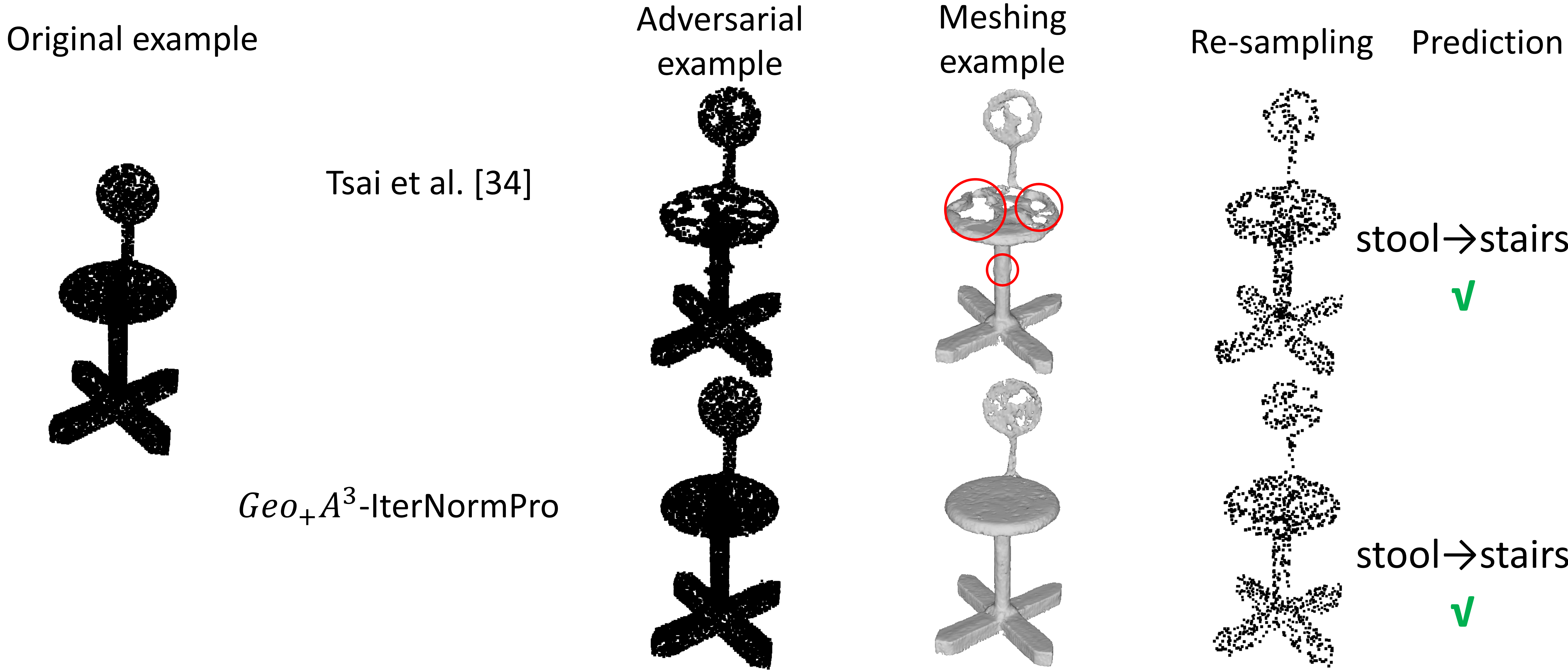}
    \end{minipage}\\
    \begin{minipage}{0.80\textwidth}
        \centering
        \includegraphics[width=0.99\textwidth]{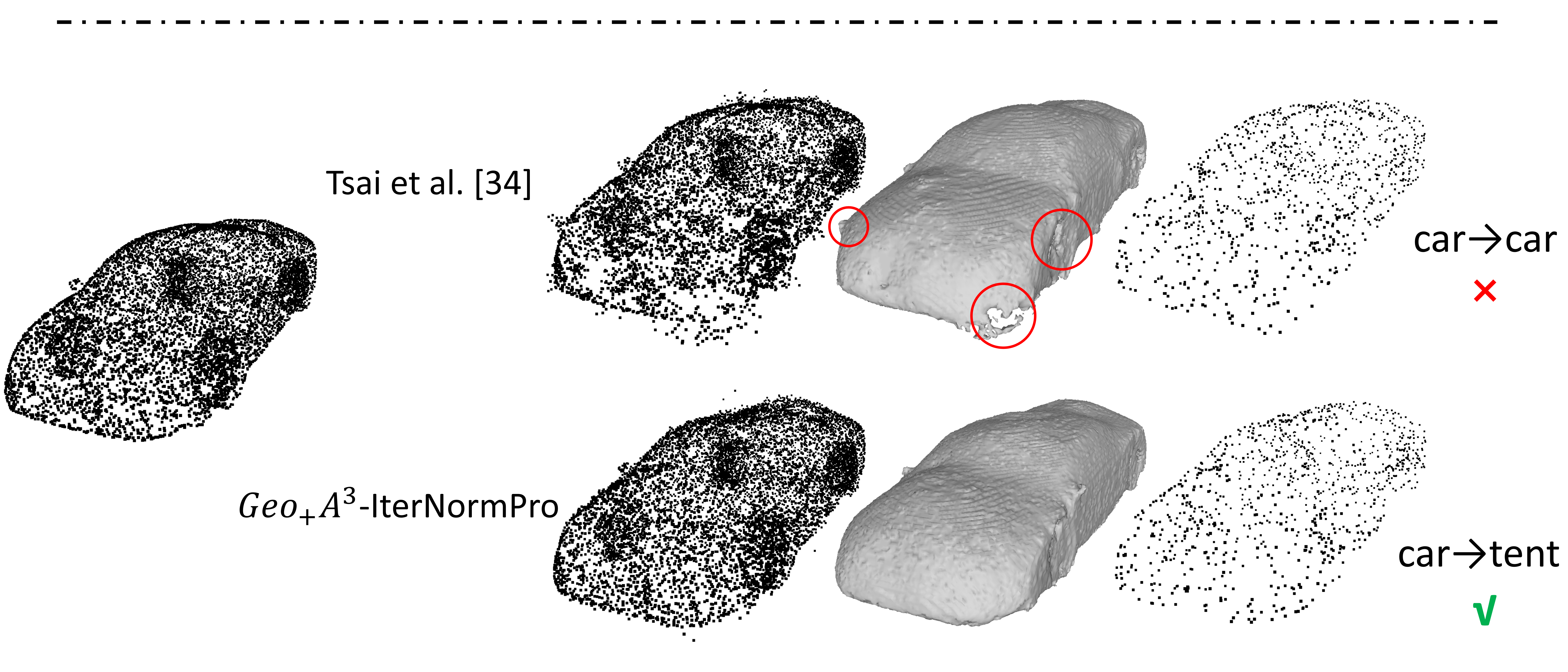}
    \end{minipage}\\
    \begin{minipage}{0.80\textwidth}
        \centering
        \includegraphics[width=0.99\textwidth]{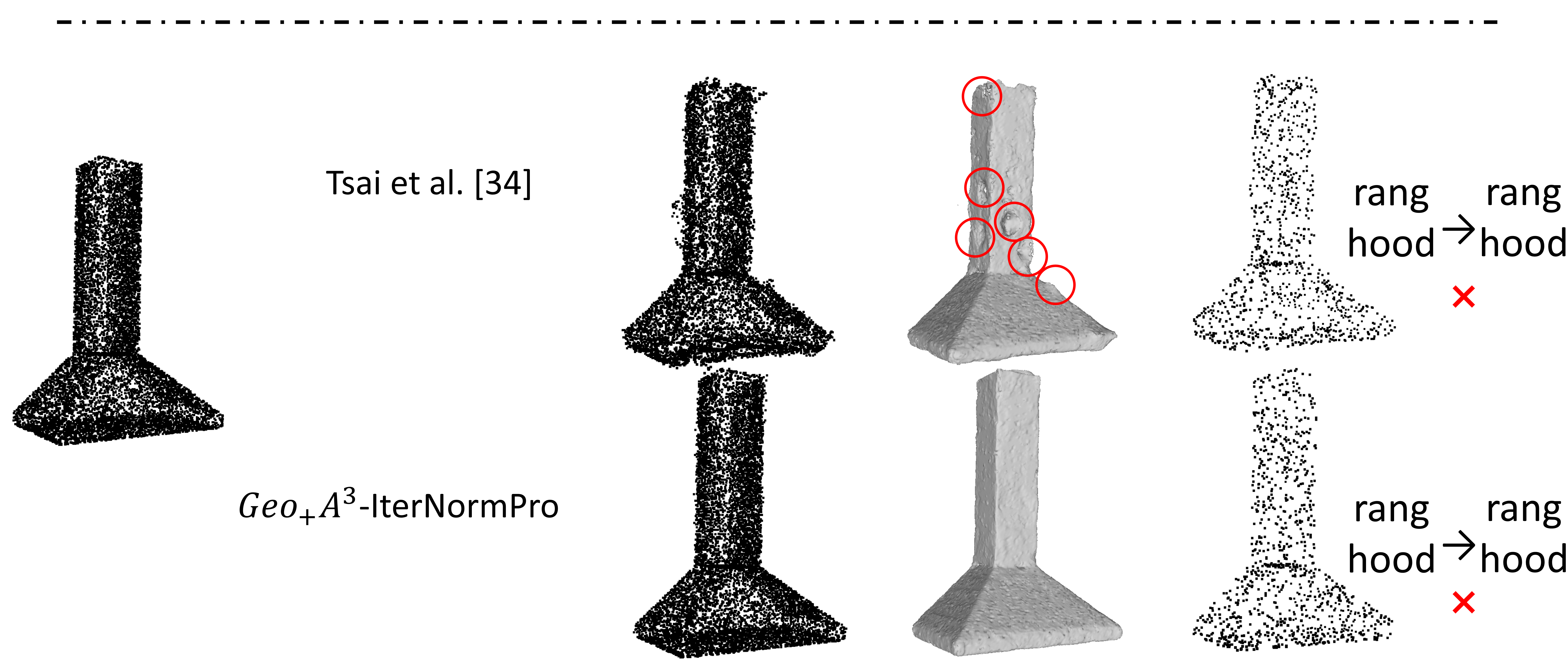}
    \end{minipage}
	\caption{
        Results of adversarial point cloud, its mesh reconstruction, and re-sampling of points from our proposed $Geo_{+}A^3$-IterNormPro and Tsai \emph{et al.} \cite{AAAICloseRelatedWork}. Red circles highlight the irregularities in the results of Tsai \emph{et al.} \cite{AAAICloseRelatedWork}. A green tick indicates a successful attack, and a red cross indicates a failed attack.
    }\label{fig:reconstruct_per}
    \vspace{-0.4cm}
\end{figure*}

\noindent\textbf{Subjective Evaluation ---}
A subjective user study is implemented on Amazon Mechanical Turk (AMT) in order to evaluate visually imperceptible quality of adversarial results.
Specifically, for each object instance, we upload onto AMT a triple of point cloud snapshots including the benign one, the adversarial one generated by Xiang \emph{et al.} \cite{CW3DAdv}, and the adversarial one generated by our $GeoA^3$; all the uploaded adversarial results attack the PointNet classifier successfully.
Online participants are asked to compare to discriminate which one of the two adversarial point clouds are visually more similar to the benign one.
To implement the subjective comparison properly, we do the following: the order of showing the two results is randomized, each participant is limited to contribute at most $30$ trials, and each adversarial result can be shown up to $50$ different participants.
In total, we collect $1,500$ trials from $128$ participants; $82.06\%$ of the trials consider adversarial results from our $GeoA^3$ visually closer to the benign ones, when compared against those from Xiang \emph{et al.} \cite{CW3DAdv}, as summarized in Table \ref{table:AMT}.
An AMT interface for our conducted subjective study is shown in Figure \ref{fig:AMT_ill}.

We finally show in Figure \ref{fig:imper_per_tenclass} more results of adversarial point clouds from our method in an adversarial attack matrix, where an example instance of each category is attacked against PointNet \cite{PointNet} targeting at all the other 9 categories.

\begin{table}[htbp]
	\caption{
        Evaluation of surface-level adversarial effects for results generated by different methods.
        For each adversarial result , we use Points2Surf reconstruction \cite{erler2020points2surf} to get its mesh $\mathcal{S}'$, from which farthest point sampling \cite{PointNetPP} is used to re-sample a $\mathcal{Q}'$.
        Performance is measured in terms of both the attack success rate (\%) and geometric regularity (\ref{Eq:GeoRegularityMeasure}) on $\mathcal{Q}'$.
		}\label{table:mesh_attack}
	\begin{center}
		\begin{small}
			\scalebox{0.87}{
				\begin{tabular}{l c c }
					\toprule
                    \multirow{2}{*}{Model} & Method & \makecell{ Attack success rate (\%) on $\mathcal{Q}'$ / \\ Geometric regularity $R(\mathcal{Q}')$ } \\
					\midrule
                    \multirow{2}{*}{PointNet} & Tsai \emph{et al.} \cite{AAAICloseRelatedWork} & $ 14.78 / 0.1031 $ \\
                    ~ & $Geo_{+}A^3$-IterNormPro & $ \bm{19.65} / \bm{0.1017} $ \\
                    \midrule
                    \multirow{2}{*}{PointNet++} & Tsai \emph{et al.} \cite{AAAICloseRelatedWork} & $ 6.24 / 0.1187 $ \\
                    ~ & $Geo_{+}A^3$-IterNormPro & $\bm{11.20} / \bm{0.1125}$ \\
                    \midrule
                    \multirow{2}{*}{DGCNN} &  Tsai \emph{et al.} \cite{AAAICloseRelatedWork} & $ 4.17 / 0.1233 $ \\
                    ~ & $Geo_{+}A^3$-IterNormPro & $ \bm{8.24} / \bm{0.1148} $ \\
					\bottomrule
				\end{tabular}}
			\end{small}
		\end{center}
\end{table}

\begin{figure}[htbp]
	\centering
	\begin{minipage}{0.42\textwidth}
		\centering
		\includegraphics[width=0.99\textwidth]{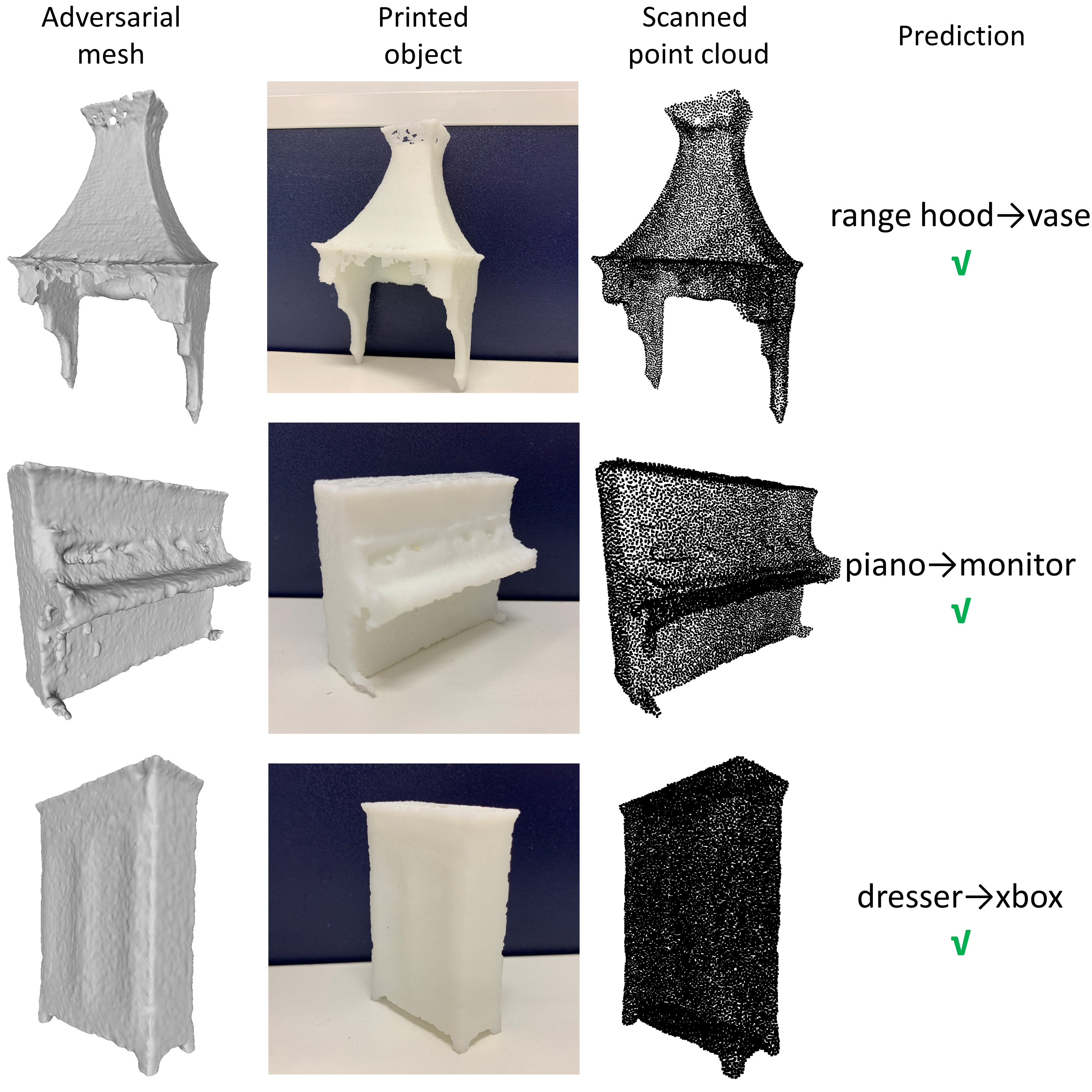}
	\end{minipage}\\
	\begin{minipage}{0.42\textwidth}
		\centering
		\includegraphics[width=0.99\textwidth]{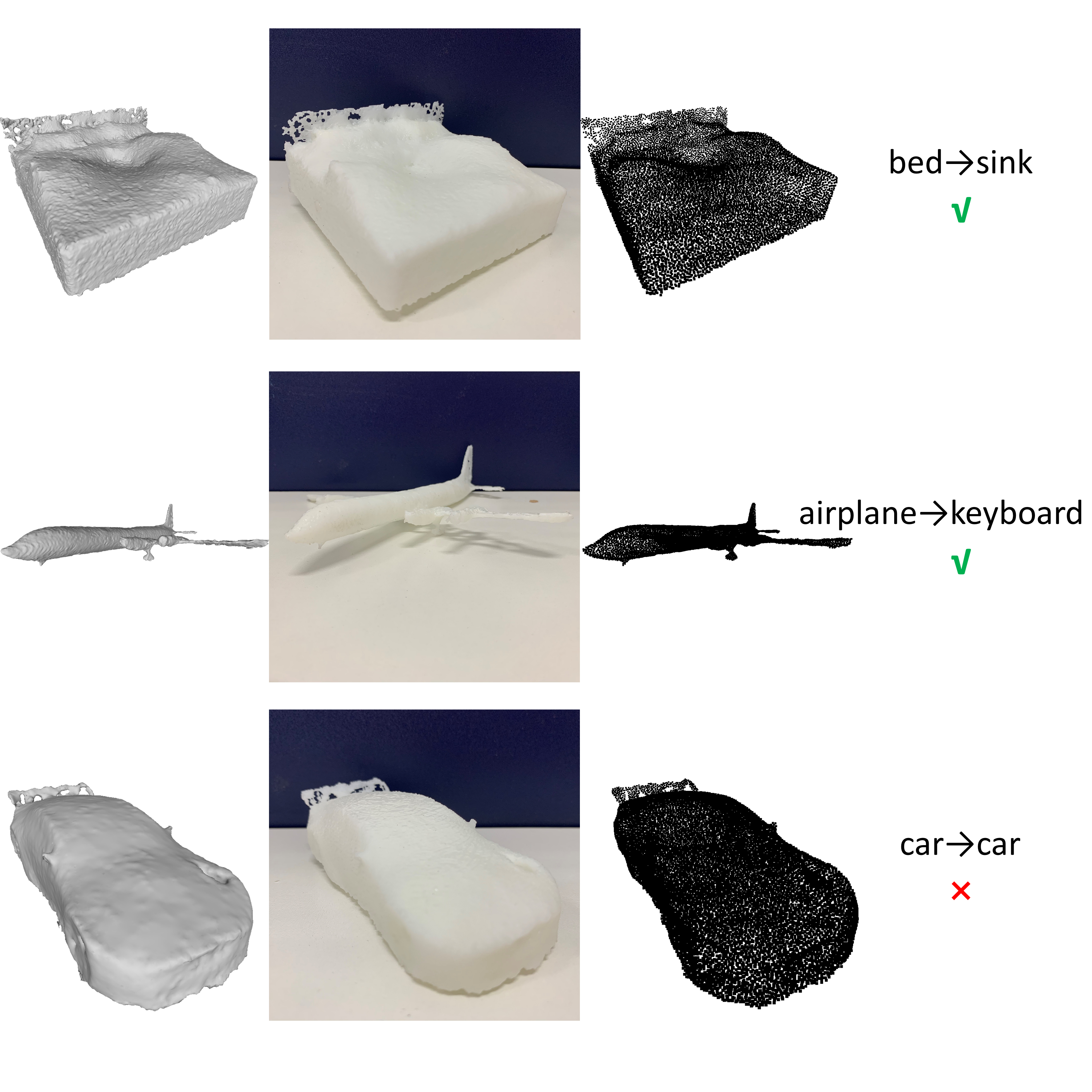}
	\end{minipage}
	\caption{
		Example meshes, their printed objects, and scanned point clouds.
		These results are obtained by first 3D-printing the mesh $\mathcal{S}'$ constructed from an adversarial $\mathcal{P}'$ generated by our $Geo_{+}A^3$-IterNormPro, and then scanning the printed object to have the point cloud $\mathcal{Q}'$.
        UnionTech Lite600HD 3D printer (materials of 9400 resin) is used for this purpose.
        Accompanying each example is the category predicted by the PointNet \cite{PointNet} model of classifier before and after attack; a green tick indicates a successful attack, and a red cross indicates a failed attack.
	}\label{fig:physical_attack}
\end{figure}

\subsection{Evaluation of Adversarial Surface Shapes and Physical Attacks} \label{expPhysicalAttack}

As discussed in Section \ref{SecPhysicalAttack}, it could be practically less defendable when adversarial effect of a generated point cloud $\mathcal{P}'$ comes more from a deformation of surface shape, which essentially defines a much challenging task of surface-level adversarial attack.
In Section \ref{SecPhysicalAttack}, we introduce a new objective function $Geo_{+}A^3$ and further propose a solving algorithm termed $Geo_{+}A^3$-IterNormPro.
To evaluate its efficacy and compare with the existing Tsai \emph{et al.} \cite{AAAICloseRelatedWork}, we follow the setting of randomly targeted attack used in \cite{AAAICloseRelatedWork}.
We use all the instances of 40 classes that are well classified in the ModelNet40 testing set.
For each instance/CAD model, we uniformly sample $10,000$ points to have the input, benign $\mathcal{P}$, whose adversarial result $\mathcal{P}'$ is obtained by $Geo_{+}A^3$-IterNormPro.
After obtaining an adversarial $\mathcal{P}'$, we use Points2Surf reconstruction \cite{erler2020points2surf} to get its mesh $\mathcal{S}'$, from which we either use farthest point sampling \cite{PointNetPP} to re-sample a point cloud $\mathcal{Q}'$, or do 3D printing using $\mathcal{S}'$ and then scan the printed object to obtain $\mathcal{Q}'$.
The respectively obtained $\mathcal{Q}'$ is used for evaluation of attacking performance.

Table \ref{table:mesh_attack} compares our method with the only existing method \cite{AAAICloseRelatedWork} that aims for better surface-level adversarial effects when generating adversarial point clouds.
For any $\mathcal{P}'$ generated by Tsai \emph{et al.} \cite{AAAICloseRelatedWork}, we do the same procedure of meshing and re-sampling to have its re-sampled $\mathcal{Q}'$.
On the three representative classification models of PointNet, PointNet++, and DGCNN, $Geo_{+}A^3$-IterNormPro consistently outperforms the Tsai \emph{et al.}.
Such results also suggest that achieving surface-level attacks via generation of adversarial point clouds is indeed a challenging task, and our contributed algorithm takes only a small step towards the desired goal. Figure \ref{fig:reconstruct_per} gives some results of the adversarial $\mathcal{P}'$, meshing $\mathcal{S}'$, and re-sampled $\mathcal{Q}'$ respectively from the two methods. In cases of either successful or failed attack, our results are smoother with less surface irregularities, while those from Tsai \emph{et al.} do have.

Among all the testing instances, we do 3D printing for 15 of them, which are selected from the successful attacking cases reported in Table \ref{table:mesh_attack}.
We do 3D printing using their respective $\mathcal{S}'$, and then scan each printed object to have the corresponding $\mathcal{Q}'$, which is used for evaluation of adversarial attack.
After 3D printing and scanning, 11 out of the 15 instances can still attack the PointNet classifier successfully, further confirming that our proposed $Geo_{+}A^3$-IterNormPro achieves a certain level of surface-level adversarial effects. Example meshes, their printed objects, and scanned point clouds are shown in Figure \ref{fig:physical_attack}.


\section{Conclusions}

In this paper, we study learning to generate adversarial point clouds in order to attack deep models of point set classifiers.
We focus on the key characterization of imperceptibility to humans that adversarial point clouds should have, which is largely overlooked in existing methods.
We analyze the different mechanisms that humans perceive 2D images and 3D shapes, and propose a new method of $GeoA^3$ whose objective combines a misclassification loss of targeted attack and a new design of geometry-aware regularizer.
Our proposed regularizer favors solutions with the desired surface properties of smoothness and fairness, while the targeted attack loss supports continuous pursuing of more malicious signals.
The combined effect enables $GeoA^3$ to generate adversarial results that are arguably less defendable and of the key adversarial characterization of being imperceptible to humans.
Comparative results confirm the advantages of our $GeoA^3$ in terms of both quantitative and qualitative measures.
To generate practically more desirable adversarial surfaces, we make an attempt in this paper and propose a simple algorithm of $Geo_{+}A^3$-IterNormPro.
Solving $Geo_{+}A^3$-IterNormPro can better preserve surface-level adversarial effects when re-sampling point clouds from the surface meshes reconstructed from the obtained adversarial point clouds.
Experiments of both synthetic and physical attacks show the efficacy of our contributed algorithm.
However, surface-level adversarial attack is still at a much lower successful rate.
In future research, we are interested in improving the attacking performance either by advanced methods of adversarial point cloud generation or by directly learning to generate adversarial meshes.

\FloatBarrier
\bibliographystyle{IEEEtran}
\bibliography{IEEEabrv,TPAMI-2020-04-0559}

\end{document}